\documentclass{article}

\usepackage{arxiv}

\usepackage[T1]{fontenc}    
\usepackage{hyperref}       
\usepackage{url}            
\usepackage{booktabs}       
\usepackage{amsfonts}       
\usepackage{nicefrac}       
\usepackage{microtype}      
\usepackage{lipsum}		
\usepackage{graphicx}
\usepackage{natbib}
\usepackage{doi}
\usepackage[most]{tcolorbox}

\usepackage[latin1]{inputenc}

\usepackage{amssymb}
\usepackage{amsmath}
\usepackage{algorithmic}
\usepackage{algorithm}
\usepackage{amssymb}

\usepackage{comment}
\usepackage{graphicx}
\usepackage{epstopdf}
\usepackage{array}
\usepackage{booktabs}
\usepackage{multirow}
\usepackage{caption}
\usepackage{subcaption}
\usepackage{tcolorbox}
\graphicspath{{figures/}}

\usepackage{listings}

\usepackage{enumitem}

\usepackage{xcolor}
\usepackage{colortbl} 
\definecolor{jade}{rgb}{0.0, 0.66, 0.42}
\definecolor{carolinablue}{rgb}{0.6, 0.73, 0.89}

\usepackage[commandnameprefix=always]{changes}
\usepackage{lipsum}
\usepackage[normalem]{ulem}
\usepackage{indentfirst}
\usepackage{tablefootnote}

\definechangesauthor[name={Chao}, color=orange]{chao}
\setaddedmarkup{\color{blue}{#1}}
\setdeletedmarkup{\color{orange}{\sout{#1}}}

\newcommand{\tabincell}[2]{\begin{tabular}{@{}#1@{}}#2\end{tabular}} 

\usepackage{url}
\makeatletter
\g@addto@macro{\UrlBreaks}{\UrlOrds}
\makeatother
\PassOptionsToPackage{hyphens}{url}

\usepackage{flushend}

\usepackage{seqsplit}

\usepackage{xspace}

\newcommand{\lbc}[1]{\COMMENT} 

\newcommand{\sysname}{MFTCoder\xspace}

\newcommand{\eg}{\textit{e.g.}\xspace}
\newcommand{\ie}{\textit{i.e.}\xspace}

\newcommand{\mft}{\textsc{MFT-5Tasks}\xspace}
\newcommand{\sftmix}{\textsc{SFT-Mixed}\xspace}
\newcommand{\sftcodecomp}{\textsc{SFT-S-CodeCompletion}\xspace}
\newcommand{\sfttextcode}{\textsc{SFT-S-Text2Code}\xspace}
\newcommand{\sftcodetrans}{\textsc{SFT-S-CodeTrans}\xspace}
\newcommand{\sftunit}{\textsc{SFT-S-UnitTest}\xspace}
\newcommand{\sftcodecomment}{\textsc{SFT-S-CodeComment}\xspace}

\usepackage[misc]{ifsym}
\usepackage{bbding}

\newcounter{finding}

\title{MFTCoder: Boosting Code LLMs with Multitask Fine-Tuning}

\author{
Bingchang Liu,
Chaoyu Chen,
Zi Gong,
Cong Liao,
Huan Wang,
Zhichao Lei,
Ming Liang, \\
\textbf{Dajun Chen},
\textbf{Min Shen},
\textbf{Hailian Zhou},
\textbf{Wei Jiang},
\textbf{Hang Yu\thanks{Corresponding Author: \texttt{\{hyu.hugo, lijg.zero\}@antgroup.com}}},
\textbf{Jianguo Li\footnotemark[1]} \\
Ant Group, China\\
}

\hypersetup{
pdftitle={MFTCoder: Boosting Code LLMs with Multitask Fine-Tuning},
pdfsubject={q-bio.NC, q-bio.QM},
pdfauthor={Bingchang Liu, Chaoyu Chen, Cong Liao, Zi Gong, Huan Wang, Zhichao Lei, Ming Liang, Dajun Chen, Min Shen, Hailian Zhou, Hang Yu, Jianguo Li},
pdfkeywords={Large Language Model, Code Generation, Multi-task Learning},
}

\begin{document}
\maketitle


\begin{abstract}
    Code LLMs have emerged as a specialized research field, with remarkable studies dedicated to enhancing model's coding capabilities through fine-tuning on pre-trained models. Previous fine-tuning approaches were typically tailored to specific downstream tasks or scenarios, which meant separate fine-tuning for each task, requiring extensive training resources and posing challenges in terms of deployment and maintenance. Furthermore, these approaches failed to leverage the inherent interconnectedness among different code-related tasks.
    To overcome these limitations, we present a multi-task fine-tuning framework, \sysname, that enables simultaneous and parallel fine-tuning on multiple tasks. By incorporating various loss functions, we effectively address common challenges in multi-task learning, such as data imbalance, varying difficulty levels, and inconsistent convergence speeds. Extensive experiments have conclusively demonstrated that our multi-task fine-tuning approach outperforms both individual fine-tuning on single tasks and fine-tuning on a mixed ensemble of tasks.
    Moreover, \sysname offers efficient training capabilities, including efficient data tokenization modes and PEFT fine-tuning, resulting in significantly improved speed compared to traditional fine-tuning methods. \sysname seamlessly integrates with several mainstream open-source LLMs, such as CodeLLama and Qwen. Leveraging the CodeLLama foundation, our \sysname fine-tuned model, \textsc{CodeFuse-CodeLLama-34B}, achieves an impressive pass@1 score of 74.4\% on the HumaneEval benchmark, surpassing GPT-4 performance (67\%, zero-shot). MFTCoder is open-sourced at \url{https://github.com/codefuse-ai/MFTCOder}
\end{abstract}

\keywords{Large Language Model \and Code Generation \and Multi-task Learning}

\section{Introduction}

The paradigm-shifting emergence of ChatGPT\footnote{https://openai.com/blog/chatgpt\label{chatgpt}}, powered by both GPT-3.5 and GPT-4~\cite{openai2023gpt4}, has set ablaze the landscape of research and development in the realm of large language models (LLMs). This breakthrough has further sparked the interest in leveraging LLMs for code understanding and generation, commonly referred to as Code LLMs. By pretraining on extensive code data sources such as the Github public data, these Code LLMs can acquire comprehensive contextual representations that can be applied to various code-related tasks~\cite{}. 

While the pretraining stage of (Code) LLMs seek to ensure their generalizability to different downstream tasks, the subsequent finetuning stage typically only adapt the (Code) LLMs to a specific task or a scenario. However, this approach overlooks two critical challenges. \textbf{Firstly, it involves resource-intensive individual finetuning of large LLMs for each task, which hinders efficient deployment in production. Secondly, the interrelated nature of code domain tasks suggests that joint finetuning can enhance performance compared to separate finetuning.} It is therefore imperative to conduct multitask finetuning, enabling simultaneous handling of all tasks while leveraging the strengths of related tasks to enhance performance. 

As an illuminating example, suppose we have two related tasks: code completion and code summarization. Code completion involves predicting the next line of code based on a partial code snippet, while code summarization aims to generate a concise human-readable summary of a given code snippet. Traditionally, separate models would be fine-tuned for each task, resulting in resource-intensive duplication. However, code completion and code summarization have inherent connections. Completion of a code snippet relies on understanding the overall functionality and purpose, while generating an accurate summary requires comprehending the structure, dependencies, and intended functionality. By employing multitask learning, a single model can be trained to jointly learn both tasks, leveraging shared knowledge and patterns, leading to improved performance on both tasks. The model understands the contextual dependencies between code elements, aiding in predicting the next snippet and generating informative summaries. Furthermore, multitask learning offers additional benefits beyond individual task performance: the shared representation between tasks helps mitigate overfitting, promote better generalization, and enhance the model's ability to handle data scarcity for specific tasks. If code completion has a larger training dataset than code summarization, the model can leverage the abundance of completion data to enhance performance in summarization, effectively addressing data scarcity challenges. Multitask learning even enables the model to handle unseen but related tasks without specific training data. Overall, multitask learning allows models to jointly learn multiple related tasks, benefiting from shared knowledge, improving performance, enhancing generalization, and handling data scarcity.

Despite the importance of multitask learning for finetuning, only a handful of existing studies have explored this approach in the domain of NLP~\cite{raffel2023exploring,aghajanyan2021muppet,aribandi2022ext5}. These studies incorporate multi-task data and merge it for large-scale model learning, without explicitly separating the tasks. Unfortunately, these studies tend to prioritize tasks with larger sample sizes, disregarding tasks with smaller sample sizes. Furthermore, they fail to ensure equal convergence speed among tasks, leading to over-optimization of some tasks and under-optimization of others.

In this paper, we focus on multitask fine-tuing (MFT) of (Code) LLMs, in order to guarantee equitable attention to tasks with varying sample sizes and approximately similar optimization progress. In particular, our attention is on Code LLMs, as code domain tasks often exhibit correlations, and so we name our approach \sysname. We emphasize that MFTcoder can be extended to an arbitrary set of related-NLP tasks in a straighforward manner. To enhance the efficiency of \sysname, we incorporate parameter-efficient fine-tuning techniques, including LoRA~\cite{hu2021lora} and QLoRA~\cite{dettmers2023qlora}. Experimental results demonstrate that multi-task models trained using the MFT approach outperform those fine-tuned individually for each task or by merging data from multiple tasks. We further validate the effectiveness of \sysname on various baseline pretrained LLMs, such as Qwen~\cite{qwen}, Baichuan~\cite{baichuan2023baichuan2}, Llama~\cite{touvron2023llama}, Llama 2~\cite{touvron2023llama2}, StarCoder~\cite{li2023starcoder}, CodeLLama~\cite{roziere2023code}, and CodeGeex2~\cite{zheng2023codegeex}. Remarkably, when applying \sysname to the CodeLlama-34B-Python~\cite{roziere2023code} base model, it achieves a pass@1 score of 74.4\% on the humanEval evaluation dataset, even surpassing the performance of GPT-4 (67\%, zero-shot)~\cite{openai2023gpt4}.

The main contributions of this paper can be summarized as follows:
\begin{itemize}[leftmargin=*]
    \item We propose \sysname, a novel multitask finetuning approach for concurrently adapting LLMs to multiple code-related tasks. Our focus is on addressing the issues of data balance and convergence speed that commonly arise in previous multitask finetuning methods.

    \item We validate \sysname on various baseline pretrained models, including Qwen~\cite{qwen}, Baichuan~\cite{baichuan2023baichuan2}, Llama~\cite{touvron2023llama}, Llama 2~\cite{touvron2023llama2}, StarCoder~\cite{li2023starcoder}, CodeLLama~\cite{roziere2023code}, CodeFuse~\cite{di2023codefuse13b}, and CodeGeex2~\cite{zheng2023codegeex}, demonstrating its compatibility with different baseline models.

    \item Extensive experiments show that the MFT approach outperforms individual fine-tuning for each task or data merging from multiple tasks. Notably, when implementing \sysname with the CodeLlama-34B-Python~\cite{roziere2023code} base model, it achieves an impressive pass@1 score of 74.4\% on the humanEval evaluation dataset, surpassing the performance of GPT-4 (67\%, zero-shot)~\cite{openai2023gpt4}. 
\end{itemize}

\section{Related Works}
\label{sec:background}

\subsection{Code LLMs}

Coding capability serves as a critical criterion for evaluating general large language models (LLMs) in code-related tasks. Notable performance on the widely-used HumanEval dataset~\cite{chen2021codex}, a benchmark for code generation, has been observed across various models, including LaMDA~\cite{thoppilan2022lamda}, PaLM~\cite{chowdhery2022palm}, PaLM 2~\cite{anil2023palm}, ChatGPT, and GPT-4~\cite{openai2023gpt4}. In particular, GPT-4 has set a remarkable record of 67.0\% pass@1 score. However, their closed-source nature limits their availability and hinders further collaborative advancements.  In contrast, recent open-source LLMs, including LLaMA~\cite{touvron2023llama}, LLaMA 2~\cite{touvron2023llama2}, Qwen~\cite{qwen}, and Phi-1.5~\cite{textbooks2}, have demonstrated notable progress in code-related tasks, with commentable scores of 23.7\%, 29.9\%, 32.3\%, and 41.4\% respectively. Despite this progress, their performance still lags behind the state-of-the-art closed-source models. 

On the other hand, LLMs specifically designed for code-related tasks, often referred to as code LLMs, have also undergone significant developments. Alongside closed-source Code LLMs such as Codex~\cite{chen2021codex}, Code-Davinci~\cite{chen2021codex}, AlphaCode~\cite{Li_2022}, PaLM-Coder~\cite{chowdhery2022palm}, and PanGu-Coder~\cite{christopoulou2022pangucoder}, open-source alternatives like including SantaCoder~\cite{allal2023santacoder}, Phi-1.0~\cite{gunasekar2023textbooks}, CodeGeeX-2~\cite{zheng2023codegeex}, StarCoder~\cite{li2023starcoder}, Code LLaMA~\cite{roziere2023code} have showcased competitive performance with their closed-source counterparts. Notably, CodeLLama-34B-Python~\cite{roziere2023code} obtains a score of 53.7\% on HumanEval. Apart from pretraining, another intriguing approach to further enhancing Code LLMs is instruction fine-tuning, as showcased by CodeT5+~\cite{wang2023codet5}, Phi-1.0~\cite{gunasekar2023textbooks}, OctoPack~\cite{muennighoff2023octopack}, and WizardCoder~\cite{luo2023wizardcoder}. By leveraging carefully curated high-quality instruction datasets, these methods exhibit the potential of fine-tuning to enhance code generation capabilities.

\subsection{Multitask Learning}

Multitask learning (MTL)~\cite{caruana1997multitask, crawshaw2020multi} is a potent approach in machine learning that holds significant promise for enhancing model performance and addressing diverse challenges~\cite{crawshaw2020multi}. By training a single model on multiple related tasks, MTL enables the model to leverage shared knowledge and patterns, leading to enhanced generalization and improved accuracy. MTL methods can be categorized into two groups: hard parameter sharing~\cite{zhao2018modulation, liu2019end, liu2019multi, kendall2018multi, liu2019loss, chen2018gradnorm, jean2019adaptive} and soft parameter sharing~\cite{duong2015low, yang2017trace, long2017learning, lee2018deep, sun2020adashare, pascal2021maximum}. Hard parameter sharing involves sharing model weights between tasks, while soft parameter sharing incorporates task-specific models with separate weights. In the context of large language models (LLMs), hard parameter sharing is particularly relevant, since the large number of parameters in LLMs facilitates their ability to handle multiple related tasks with a common set of parameters. As a result, optimizing an LLM to effectively tackle multiple tasks lies at the heart of MTL for LLMs. In recent years, notable advancements have been made in MTL techniques. Google introduced T5~\cite{raffel2023exploring} in 2020, where they explored the application of MTL techniques. Meta further introduced Mupper~\cite{aghajanyan2021muppet} in 2021, which applies multi-task learning between pretraining and fine-tuning, termed as pre-fine-tuning (PFT). They discovered that incorporating this step enhances the performance of the pretrained model across various downstream tasks and significantly improves the speed of fine-tuning. However, if the number of tasks in PFT is too small, it can have a negative impact on the model's performance. Therefore, it is recommended to have a minimum of 15 tasks for optimal results. Building upon T5, Google introduced ExT5~\cite{aribandi2022ext5}, which increased the number of tasks to 107. They found that as long as the number of tasks in pretraining is sufficiently large, even if there may be mutual interference among tasks, the ultimate results are still remarkably good. Ultimately, ExT5 outperformed T5 across multiple metrics. It is worth noting that these studies mainly focused on incorporating multi-task data and merging it for the large model to learn, without explicitly segregating the tasks. While these approaches have shown promising results, they tend to overlook data imbalance and convergence speed issues that often arise in MTL. In this paper, we address these challenges and propose \sysname, a multitask finetuning approach for LLMs that tackles these problems effectively.

\section{Approach}

\begin{figure}[!t]
    \centering
    \includegraphics[width=\textwidth]{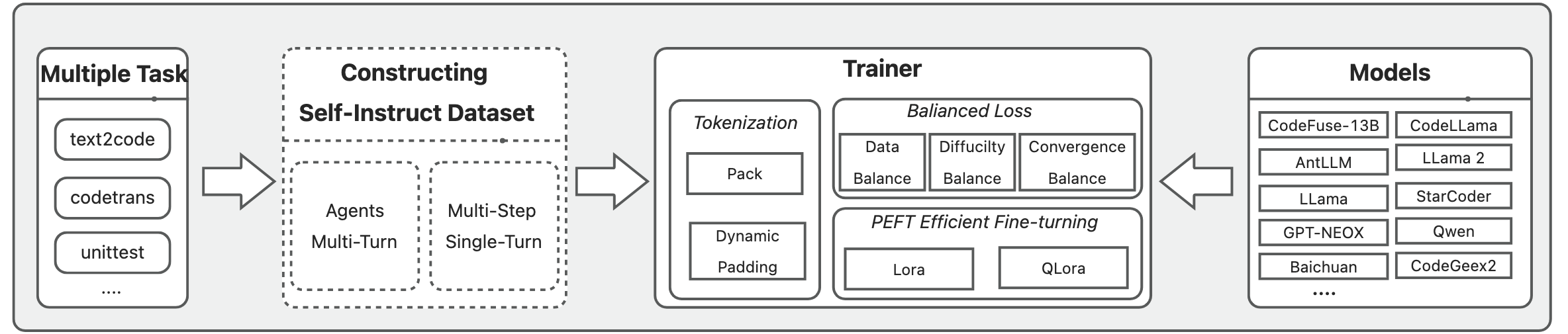}
    \caption{
      Overview of \sysname framework.}
    \label{fig:framwork}
\end{figure}

In this section, we will introduce our multi-task fine-tuning framework, \sysname~\footnote{\label{mftcoder}https://github.com/codefuse-ai/MFTCoder}, along with the design of its key components.

\subsection{MFT Framework}

\sysname aims to seamlessly adapt LLMs to diverse new scenarios while maximizing their performance within a specific context. When applying \sysname to a new scenario, the initial step involves decomposing the scenario into smaller tasks that correspond to targeted abilities. For instance, in the domain of code LLMs, the overarching objective of enhancing models' code capabilities can be further divided into specific tasks like code completion, text-to-code generation, unit test case generation, code repair, code debugging, and even cross-language translation. Our extensive practical experience has demonstrated that \sysname effectively handles multi-task scales ranging from single to dozens or even hundreds of tasks.
Each task necessitates the collection and organization of fine-tuning datasets. However, data collection for certain tasks can pose challenges. To overcome this, \sysname leverages Self-Instruct~\cite{wang2022self} techniques and Agents to generate instruction datasets.
With the capability to concurrently fine-tune multiple downstream tasks, \sysname effectively handles substantial volumes of fine-tuning data, ensuring efficient training. It incorporates two efficient data tokenization modes and implements PEFT (Parameter-Efficient Fine-Tuning) techniques to enhance training efficiency.
In the realm of multi-task learning, \sysname confronts the issue of task imbalances, encompassing imbalanced data distribution, varying task difficulties, and divergent convergence rates. To mitigate these challenges, \sysname introduces or adapts different loss functions to achieve task balance.
Recognizing that different large-scale models possess distinct strengths and capabilities, \sysname facilitates the selection of suitable model architectures based on specific scenarios to achieve optimal performance. It has been adapted to popular LLMs, including LLama~\cite{touvron2023llama}, LLama 2~\cite{touvron2023llama2}, CodeLLama~\cite{roziere2023code}, Qwen~\cite{qwen}, Baichuan 1/2~\cite{baichuan2023baichuan2}, ChatGLM 2~\cite{du2022glm}, CodeGeeX 2~\cite{zheng2023codegeex}, GPT-NEOX~\cite{black2022gptneox20b}, CodeFuse~\cite{di2023codefuse13b}, StarCoder~\cite{li2023starcoder}, AntLLM, and more. We continuously update and expand the compatibility with additional models.

The overall framework of \sysname is illustrated in Figure~\ref{fig:framwork}. In the subsequent sections, we will provide a more detailed exploration of these components, including instruction datasets construction, efficient tokenization modes, PEFT fine-tuning and balanced loss functions.

\subsection{Instruction Dataset Construction}



\begin{figure}[!t]
    \centering
    \includegraphics[width=.8\textwidth]{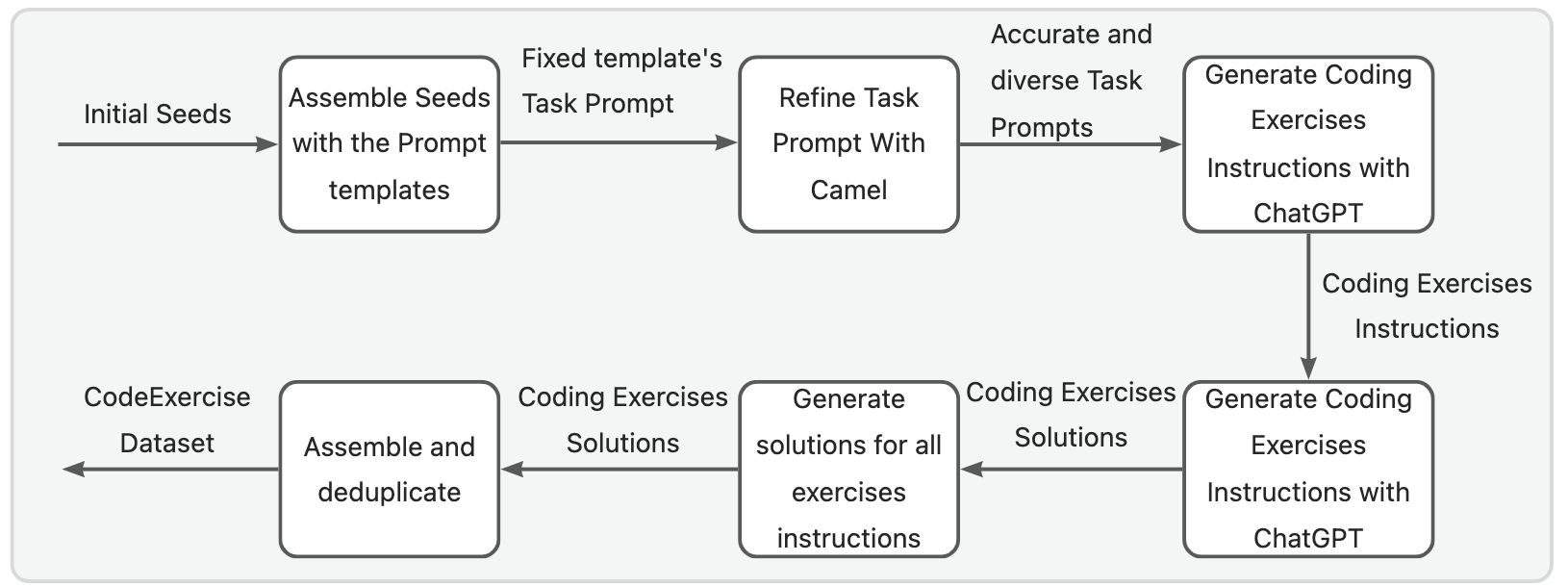}
    \caption{
       Data Generation Approach for Code Exercises Datasets using Single-turn Conversation Scheme.}
    \label{fig:single_turn_data_generation}
\end{figure}

For tasks with challenging data collection, We employ the Self-Instruct~\cite{wang2022self} technique to generate fine-tuning data for downstream code-related tasks in \sysname. This involves providing customized prompts to GPT-3.5 or GPT-4 that clearly describe our instruction generation requirements, thereby generating instructional data.
Furthermore, we drew inspiration from the Textbook approach employed in the PHI work~\cite{gunasekar2023textbooks}, incorporating the self-instruct technique to generate Code Exercises datasets for downstream code-related tasks.

In terms of specific implementation, we have two options. One is the Agents multi-turn conversation approach achieved through Camel~\cite{li2023camel}, and the other is the single-turn conversation method by directly invoking the ChatGPT API. 
In our multi-turn approach, we employ Camel to launch two agents, each assigned specific roles and task themes, facilitating a dialogue between them to generate instructional data aligned with the given theme. For instance, when generating Python exercise data, we designate the roles of 'teacher' (simulating the user role of ChatGPT) and 'student' (simulating the assistant role of ChatGPT) for the agents. The teacher's responsibility is to provide exercise instructions to the student, while the student's task is to offer corresponding solutions to those instructions. This iterative process continues, generating multiple exercise questions, until the task requirements are met or the maximum input length of ChatGPT is reached. To accommodate ChatGPT's input length limitation, we cannot directly utilize a large question as the task theme. For instance, when creating Python exercise questions to assess students' proficiency, we break down the main theme into smaller Python knowledge points (\eg binary search tree) and initiate separate Camel sessions for each knowledge point. For a concrete example, please refer to Appendix~\ref{apx:camel}.

The multi-turn approach provides high automation but can be costly due to the need for maintaining two agents, each making multi-turn calls to the ChatGPT API. To mitigate this, we propose a more cost-effective single-turn conversation generation approach, and the overall process is illustrated in Figure~\ref{fig:single_turn_data_generation}.
We begin by creating an initial set of seeds, such as hundreds of Python knowledge points. These seeds are then combined with prepared fixed prompt templates to generate a set of patterned task prompts. To address the issue of reduced diversity caused by fixed templates and to ensure accurate prompt descriptions, we utilize Camel's task prompt refinement feature to obtain precise and diverse task prompts. Each task prompt is used to generate a set of instructions related to the corresponding seed (\eg exercise problems related to binary search trees). Using ChatGPT, we generate the corresponding solutions for the generated instructions. Finally, we assemble and deduplicate the instructions with their respective solutions to obtain an exercise dataset.
We have open-sourced a Python Code Exercises dataset~\footnote{https://huggingface.co/datasets/codefuse-ai/CodeExercise-Python-27k} constructed using this approach.

\subsection{Efficient Tokenization Modes}

\begin{figure}
    \centering
    \begin{subfigure}{0.6\textwidth}
        \centering
        \includegraphics[width=\linewidth]{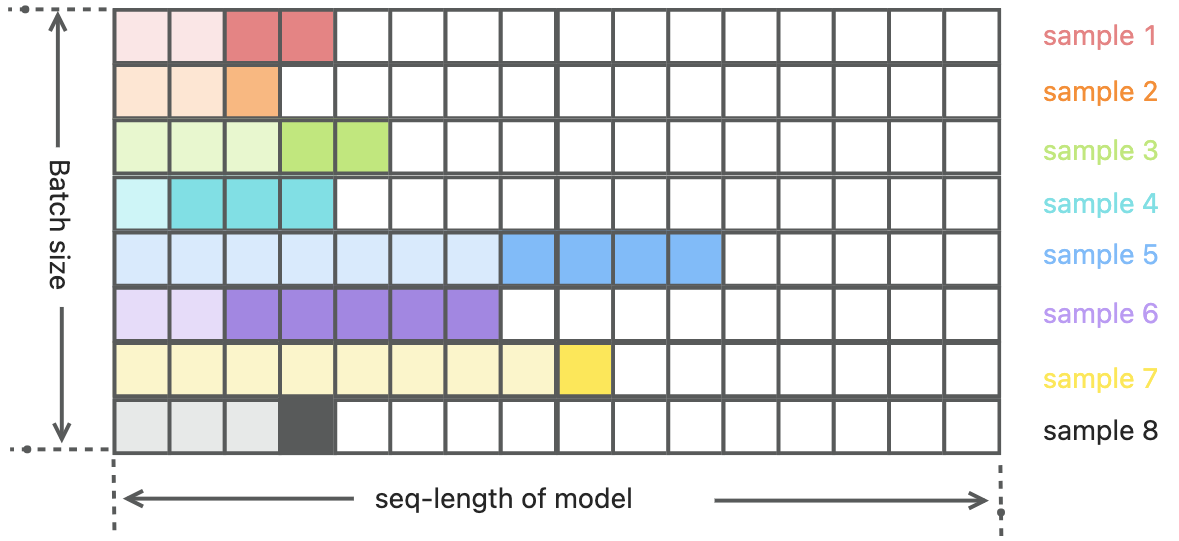}
        \caption{Normal SFT Mode}
        \label{fig:normal_sft_data}
    \end{subfigure}
    \vskip\baselineskip
    \begin{subfigure}{0.45\textwidth}
        \centering
        \includegraphics[width=\linewidth]{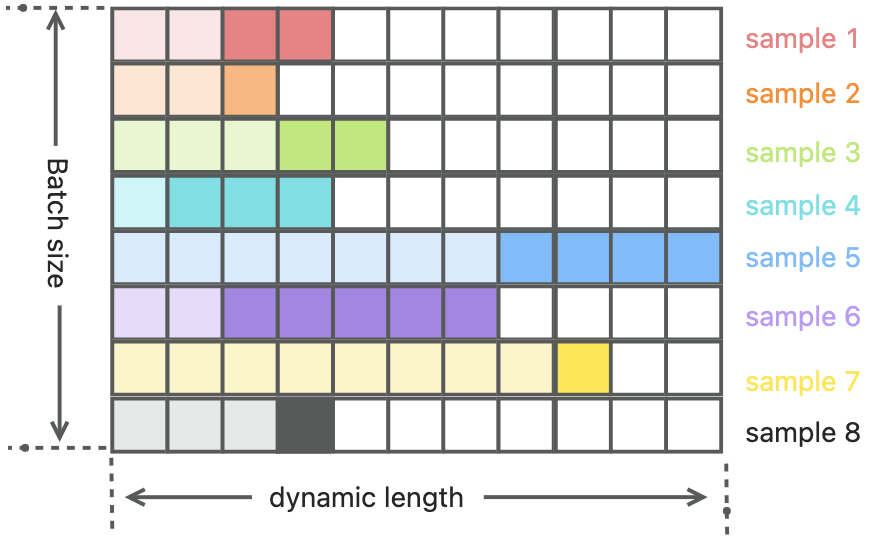}
        \caption{Dynamic Padding Mode}
        \label{fig:dynamic_padding}
    \end{subfigure}
    \hfill    
    \begin{subfigure}{0.45\textwidth}
        \centering
        \includegraphics[width=\linewidth]{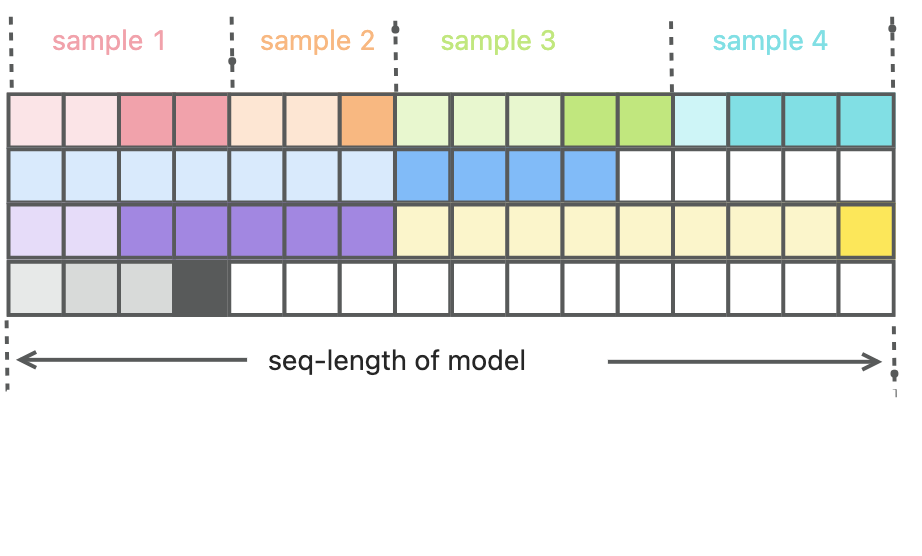}
        \caption{Pack SFT Mode}
        \label{fig:pack_sft_data}
    \end{subfigure}
    \caption{Illustration of the differences in sample organization within a batch between normal SFT, dynmaic padding and Pack SFT tokenization modes. The light-colored squares in the figure represent the Prompt section of the samples, while the dark-colored squares represent the Label section (participating in loss calculation). The blank squares represent padding section.}
    \label{fig:sft_pack}
\end{figure}

Tokenization is an essential step in the pre-training and fine-tuning of LLM models, where input and output texts are split into smaller units to be processed. It, along with the loss function, effectively defines how the data is utilized during the training process, thus playing a crucial role in both the model's effectiveness and training efficiency. 
In the typical SFT tokenization scheme, samples within the same batch are aligned to the maximum input length (seq-length) of the model with extra padding tokens, shown as Figure~\ref{fig:normal_sft_data}. However, in practice, we have found that this approach results in a high proportion of padding tokens. For example, when using the CodeFuse-13B~\cite{di2023codefuse13b} tokenizer to process 35 downstream tasks, the average proportion of padding tokens is 92.22\% (with seq-length set to 4096). This means a significant number of tokens are used solely for alignment purposes, providing no value to the training process. This results in lower training efficiency and wastage of offline tokenization storage space. To address this issue, we have adopted and optimized two tokenization modes, namely \textit{dynamic padding} and \textit{pack} modes.

In dynamic padding mode, the micro batch window size of each GPU is determined by the maximum sample length in the micro batch. Shorter samples are padded with additional padding tokens to match this size, as shown in Figure~\ref{fig:dynamic_padding}. Although padding tokens do not affect the model's training effectiveness, they add computational overhead during training, impacting the training speed. Dynamic padding mode effectively reduces the proportion of padding tokens used, leading to faster training. In our experience, this approach can yield approximately a twofold speed improvement compared to the traditional SFT tokenization mode (actual improvement depends on the dataset). It's important to note that this mode is suitable for online tokenization scenarios only.

While the dynamic padding mode reduces the micro batch window size, the pack mode, similar to Llama 2's SFT tokenization mode~\cite{touvron2023llama2}, maximizes the utilization of the model's maximum input window length (seq-length). In the pack mode, multiple fine-tuning samples are sequentially packed into a window of seq-length, separated by eos tokens, as shown in Figure~\ref{fig:pack_sft_data}. In the figure, samples 1-4 of Figure~\ref{fig:normal_sft_data} are combined and placed in one window after one another.
If a sample cannot fit in the current window, it is placed in the next window with padding tokens filling the remaining space. For instance, in Figure~\ref{fig:pack_sft_data}, sample 5 is placed in the second window with padding tokens, while sample 6 is accommodated in the third window. 
The pack mode, in comparison to the dynamic padding mode, offers even lower padding token ratio, resulting in improved training speed. Our practical experience demonstrates that this approach reduces the average proportion of padding tokens to less than 10\% among the 35 tasks mentioned earlier, leading to a substantial boost in training speed while maintaining training effectiveness. It is important to highlight that \sysname supports both online and offline pack tokenization scenarios, serving not only the SFT phase but also the pre-training stages.


\subsection{PEFT Efficient Fine-tuning}

The prevalent large-scale models typically contain billions of parameters, while multi-task learning scenarios often involve numerous tasks, resulting in a substantial total number of fine-tuning samples. If we were to opt for full-fledged fine-tuning of these large models using a vast amount of data, two challenges would arise: firstly, the need for extensive storage and computational resources; secondly, the potential risk of catastrophic forgetting during training. To address these issues, \sysname incorporates the PEFT (Parameter-efficient fine-tuning) technique~\cite{houlsby2019parameterefficient}, enabling efficient fine-tuning to be accomplished within a short timeframe and with minimal resource requirements.

\begin{figure}[!t]
    \centering
    \includegraphics[width=.4\textwidth]{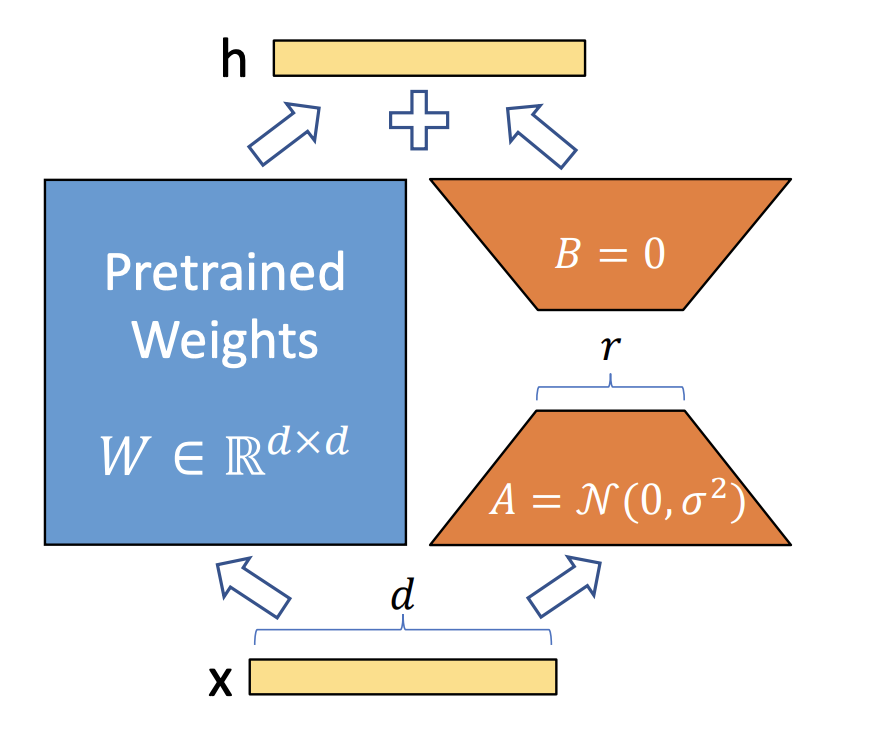}
    \caption{
       Visualizing the Essence of Lora's Basic Idea.}
    \label{fig:lora}
\end{figure}

Specifically, \sysname supports two PEFT methods: Lora (Large-scale Language Model Low-Rank Adaptation)~\cite{hu2021lora} and QLora (Quantized Large-scale Language Model Low-Rank Adaptation)~\cite{dettmers2023qlora}. The fundamental concept of Lora is quite simple, as depicted in Figure~\ref{fig:lora}. It involves adding an auxiliary branch to the original pretrained language model. During training, the parameters $W \in \mathbb{R}^{d \times d}$ of the original pretrained model remain fixed, while only the dimensional expansion matrix $A \in \mathbb{R}^{d \times r}$ and dimensional reduction matrix $B \in \mathbb{R}^{r \times d}$ within the auxiliary branch are trained. The matrix product $BA$ is then added to the original model $W$, resulting in the newly trained model.
Due to the significantly smaller magnitude of r compared to d, the number of trainable parameters can be dramatically reduced.
Building upon LoRA, QLoRA incorporates a novel high-precision quantization technique called NF4 and dual quantization to quantize the pretrained model to 4 bits. It also introduces a small set of learnable low-rank adapter weights. These weights are fine-tuned by optimizing the gradients through back-propagation of the quantized weights. As a result, QLoRA enables the fine-tuning of larger models using fewer GPU resources.
As an example, \sysname can fine-tune a 70B model on a single A100 with 80GB of VRAM.

\subsection{Multitask Fine-Tuning with Balanced Losses}


As a multi-task learning framework, \sysname, as described in Section~\ref{sec:background}, faces a significant challenge of data imbalance, task heterogeneity, and varying convergence speeds. To address these challenges, \sysname incorporates a set of loss functions specifically designed to alleviate these imbalances.



To address the issue of data imbalance, we first ensure that all samples from all tasks are utilized exactly once within a single epoch. To avoid the model favoring tasks with larger amounts of data, we introduce a weight assignment strategy during loss computation. Specifically, we support two weight calculation schemes: one based on the number of task samples and the other based on the number of valid tokens involved in the loss calculation. The former is more straightforward, but it may perform poorly when dealing with tasks that have extreme differences in the number of valid tokens, such as binary classification tasks like "yes" or "no" answering or single-choice exam tasks. On the other hand, the latter weight assignment scheme based on the actual number of valid tokens involved in the loss calculation can mitigate these issues. The specific formulation for weighted loss calculation is shown in Equation~\ref{eq:case3_loss}.
In Equation~\ref{eq:case3_loss}, $N$ represents the total number of tasks, $M_i$ denotes the number of samples for the i-th task, $T_ij$ signifies the count of valid tokens (i.e., tokens involved in loss calculation) for the j-th sample of the i-th task, and $t_{ijk}$ refers to the k-th valid token of the j-th sample for the i-th task.

\begin{equation}
\label{eq:case3_loss}
    \mathcal{L}(\theta)=\min\limits_{\theta}
\frac{1}{N}\sum_{i=1}^N \frac{\sum_{j=1}^{M_i} \sum_{k=1}^{T_{ij}}-\log(p_{\theta}(t_{ijk}))}{\sum_{j=1}^{M_i} T_{ij}} 
\end{equation}

To address the issue of task heterogeneity, we drew inspiration from the focal loss approach and incorporated it into \sysname. We implemented two different levels of focal loss functions to cater to different granularities. One operates at the sample level, as shown in Equation~\ref{eq:focal_loss_sample}, while the other operates at the task level, as shown in Equation~\ref{eq:focal_loss_task}.

\begin{equation}
\label{eq:focal_loss_sample}
    \mathcal{L}_2(\theta) = \min\limits_{\theta}\frac{\sum^{N}_{i=1}\sum^{M_i}_{j=1}-\alpha_i * (1 - P_{ij}) ^ {\gamma} * Q_{ij}}{\sum^{N}_{i=1}M_i},\\
    P_{ij} = \frac{1}{T_{ij}}\sum^{T_{ij}}_{k=1} P_{ijk} ,\\
    Q_{ij} = \frac{1}{T_{ij}}\sum^{T_{ij}}_{k=1} \log(P_{ijk})
\end{equation}

\begin{equation}
\label{eq:focal_loss_task}
    \mathcal{L}_3(\theta) = \min\limits_{\theta}\frac{1}{N}\sum^{N}_{i=1}-\alpha_i * (1 - P_i) ^ {\gamma} * Q_i, \\
    P_i = \frac{1}{M_i}\sum^{M_i}_{j=1}\frac{1}{T_{ij}}\sum^{T_{ij}}_{k=1} P_{ijk}, \\
    Q_i = \frac{1}{M_i}\sum^{M_i}_{j=1}\frac{1}{T_{ij}}\sum^{T_{ij}}_{k=1} \log(P_{ijk})
\end{equation}

To address the issue of inconsistent convergence speeds, we drew inspiration from the FAMO~\cite{liu2023famo} approach and innovatively applied it to calculate the validation loss. Firstly, we assumed that each task, indexed by $i$, has its own original loss $\mathcal L^i(\theta)$. In the $t$-th iteration, we updated the weights of each task based on the gradients of their corresponding validation losses, aiming to maximize the weight $w^i$ for the task with the slowest convergence speed, shown as Equation~\ref{eq:famo_loss}. Here, $g_t$ represents the gradient of the weighted validation loss for all tasks, $c^i(\alpha,g_t)$ denotes the slope (gradient) of the validation loss for the $i$-th task, $\theta_t$ denotes the parameters of the network in the $t$-th iteration, $\alpha$ is the learning rate, and $\epsilon$ is a small constant to prevent division by zero.
Furthermore, we would like to provide further explanation on how we achieve balanced convergence. To ensure that tasks converge at a similar pace, we introduce a dynamic balancing mechanism. At each iteration, we update the task-specific weights based on the gradients of their validation losses. This approach aims to give more importance to tasks with slower convergence speeds, allowing them to have a larger influence on the overall optimization process. By dynamically adjusting the task weights, we create a balanced convergence scenario, where all tasks progress towards their optimal solutions at a similar rate. This mechanism effectively addresses the issue of disparate convergence speeds and enhances the overall stability and performance of the \sysname framework.

\begin{equation}
\label{eq:famo_loss}
    \mathcal{L}_4(\theta) = \max_{g_t}\min_{i} \frac{1}{\alpha}c^i(\alpha,g_t) - \frac{1}{2}\|g_t\|^2, \\
    g_t = \sum_i w_t^i\nabla\mathcal L^i(\theta_t), \\
    c^i(\alpha,g_t) = \frac{\mathcal L^i(\theta_t) - \mathcal L^i(\theta_t - \alpha d_t)}{\mathcal L^i(\theta_t) + \epsilon}
\end{equation}

By incorporating these different loss functions, \sysname effectively addresses the diverse requirements of various multitask scenarios and alleviates the challenges of data imbalance, task heterogeneity, and inconsistent convergence speeds typically encountered in existing large-scale MTL research. \sysname's flexible framework provides a robust solution to these issues, empowering the development of more efficient and accurate multitask models.

\section{Evaluation}


In this section, we will conduct multiple sets of experiments using MFTCoder to validate the effectiveness and superiority of the MFT method. Specifically, we aim to address the following three research questions:
\begin{enumerate}[label=RQ\arabic*:, wide=4pt, leftmargin=*]
    \item {\bf Does the MFT model, obtained by fine-tuning multiple tasks using MFT methodology, outperform the SFT-S(ingle) models, where each task is individually fine-tuned?} 
    
    \item {\bf Does the MFT model outperform the SFT-Mixed model, where multiple tasks are combined and fine-tuned as one?} 
    
    \item {\bf In terms of generalization to unseen tasks, does the MFT model outperform the SFT-Mixed model?} 
\end{enumerate}

Next, we will commence by presenting the experimental setup. Subsequently, we will showcase and delve into the experimental results. Finally, we will culminate by summarizing and addressing the research questions raised in this section.

\subsection{Evaluation Setup}

To address these three research questions, we selected 5 code-related downstream tasks and prepared the corresponding fine-tuning data, as shown in Table~\ref{tab:traing_data}. Table~\ref{tab:traing_data} presents the desired enhancements (Column \textrm{III}) and the number of samples (Column \textrm{IV}) for each task. For instance, the \textsc{codecompletion-task} aims to improve the model's code completion ability and includes 192,547 fine-tuning samples. The \textsc{codetrans-task} aims to enhance the model's code translation capability and consists of 307,585 fine-tuning samples.
Thus, we trained 7 models (Column \textrm{I}), including individual \textsc{SFT-S-*} models trained for each downstream task, a combined \textsc{SFT-Mixed} model for the 5 task data, and an \textsc{MFT-5Tasks} model trained using the MFT method.

\begin{table}[!t]
  \caption{Various experimental models and their corresponding training data.}  
  \centering
  \label{tab:traing_data}
  \begin{tabular}{lcccccc}
  \toprule
   \textbf{Experimental Model} & \textbf{Task}    &  \textbf{Desired Ability}    &  \textbf{\#Samples}  & \textbf{\tabincell{c}{\#Samples \\ after packing}}   \\
  \midrule
  \sftcodecomp &  \textsc{code-completion} & Code Completion & 192,547 & 18,811  \\
  \sfttextcode &  \textsc{text2code}  &  Text-to-code Generation  &  94,086  & 14,399  \\
  \sftcodecomment & \textsc{code-comment} &  Code Comments Generation & 645,711 
 & 134,775   \\
  \sftcodetrans & \textsc{code-trans}  & Code Translation & 307,585 & 71,573  \\
  \sftunit & \textsc{unit-test} & Unit test-case generation & 390,393 & 77,681  \\
  \midrule
  \sftmix & Mix of the above 5 tasks & All of the above & 1,630,322 & 317,239 \\
  \midrule
  \mft  &  The above 5 tasks   & All of the above  &   1,630,322   & 317,239  \\
  \bottomrule
  \end{tabular}
\end{table}

In the experiment, all models were configured identically except for the training data. The base model for all models was CodeLlama-13B-Python~\cite{roziere2023code}. Each model was trained using 16 A100 GPUs (with 80GB VRAM), a micro batch size of 8, and a global batch size of 128. The Adam optimizer~\cite{kingma2017adam} was used with an initial learning rate of 2e-4, and a minimum learning rate of 1e-5. We employed the QLora-INT4 mode of \sysname for fine-tuning, with a consistent fine-tuning parameter proportion of 2.52\%. The positions and initial values of the trainable parameters were also the same. 
All models incorporate the Data-Balance Loss (i.e., Equation~\ref{eq:case3_loss}) and employ pack tokenization mode. 
Notably, when there is only one task, this loss function aligns with the conventional loss employed in standard GPT model pre-training.
To determine the convergence point for each model, we trained them until the validation loss surpassed the loss from the current epoch for the next two consecutive epochs. This mechanism, known as early-stopping strategy, was employed to ensure optimal convergence for each model.

\subsection{Evaluation Datasets}
\label{sec:eval_datasets}

In this paper, we utilized publicly available and representative code assessment benchmarks for comparative evaluation. These benchmarks include:
\begin{itemize}[leftmargin=*]
    \item \textbf{HumanEval}~\cite{chen2021codex} is a widely used Python code completion evaluation dataset, meticulously curated by researchers at OpenAI.

    \item \textbf{HumanEval-X}~\cite{zheng2023codegeex} is an extension of HumanEval, translated into multiple programming languages, enabling multi-language code completion evaluation.

    \item \textbf{DS-1000}~\cite{Lai2022DS1000} focuses on assessing a model's ability to perform data science analysis using Python code, covering essential libraries such as Numpy, Pandas, TensorFlow, Pytorch, Scipy, Sklearn, and Matplotlib.

    \item \textbf{MBPP}~\cite{austin2021program} comprises 1000 Python programming problems, constructed through crowdsourcing, primarily targeting a model's proficiency in basic Python. In this study, we selected 500 problems with ID 11-510 from MBPP to evaluate the text-to-code generation capability, specifically generating code based on problem descriptions.

    \item \textbf{CodeFuseEval}~\cite{di2023codefuse13b}, building upon HumanEval and HumanEval-X,  further extends the evaluation to include Chinese code completion (with Chinese docstrings), code translation, and unit test case generation capabilities, referred to as \textbf{CodeFuseEval-CN}, \textbf{CodeFuseEval-CodeTrans}, and \textbf{CodeFuseEval-UnitTest}, respectively.
\end{itemize}
Throughout these evaluation datasets, we employed "pass@1" as the evaluation metric in this paper.




    

  

\subsection{Evaluation Results}\label{sec:eval_results}

In this section, we will showcase the evaluation results of seven trained models. 
For the \textsc{SFT-S-*} models, which were trained individually for each task, we will focus on testing their specific target capabilities. For instance, we will exclusively evaluate the performance of the \sftcodecomp model in the code completion task. On the other hand, for the \sftmix and \mft models, we will assess their performance on each task and compare it with the corresponding \textsc{SFT-S-*} models.
Specifically, we will conduct tests to evaluate the capabilities of code completion, text-to-code generation, code comment generation, code translation, and unit test case generation.

\subsubsection{Code Completion}

For code completion, we employed the HumanEval~\cite{chen2021codex} and HumanEval-X ~\cite{zheng2023codegeex} evaluation datasets to assess the model's performance. HumanEval is a widely-used benchmark dataset released by OpenAI specifically designed to evaluate the Python code completion ability of large language models. HumanEval-X, on the other hand, is an expansion of HumanEval that enables the evaluation of large models' code completion performance across various programming languages. Consistent with other studies, we employed the pass@1 metric as the evaluation measure. 

 We evaluated three models: \sftcodecomp, \sftmix, and \mft. The performance of these models on the HumanEval dataset is summarized in Table~\ref{tab:humaneval_mbpp} (Column \textrm{III}). Results indicate that the \mft model, trained using the MFT approach, outperforms the other two models. It achieves a 2.44\% higher performance compared to the \sftmix model, which was fine-tuned with mixed task data. It is worth noting that the \sftmix model does not perform as well as the \sftcodecomp model, which was trained individually for the code-completion task.

 Furthermore, we conducted a multilingual evaluation on the HumanEval-X dataset for the three models, as presented in Table~\ref{tab:humaneval_x}. The \mft model demonstrates superior performance in Java and Golang, while the \sftmix model excels in C++ and JavaScript. Overall, the evaluation affirms that the \mft model outperforms the others, with an average improvement of 1.22\% over the \sftmix model.

 \textbf{Overall, in terms of code completion tasks, models trained using the MFT method outperform both individually fine-tuned models and models fine-tuned after combining multiple tasks.}

\begin{table}[!t]
  \caption{Pass@1 performance on HumanEval (Code Completion) and MBPP (Text-to-Code Generation). We utilized the greedy decoding strategy with zero-shot. The values of CodeLlama-Python-base are taken from~\cite{roziere2023code}.}  
  \centering
  \label{tab:humaneval_mbpp}
  \begin{tabular}{lcccc}
  \toprule
   \textbf{Model} &  \textbf{Size} &  \textbf{\tabincell{c} {Humaneval \\ pass@1}}  & \textbf{\tabincell{c}{MBPP \\ pass@1}}  & \textbf{Average} \\ 
  \midrule

 CodeLlama-Python-base~\cite{roziere2023code} & 13B & 43.3\% & 49.0\%  & 46.15\% \\
 \hline

 \sftcodecomp & 13B & 59.76\% & NA & NA \\

 \sfttextcode & 13B & NA & 54.2\%  & NA \\

 \sftmix & 13B &  57.93\% & 53.6\% & 55.765\% \\

\rowcolor{gray!50}
 \mft & 13B  & \textbf{60.37\% }& \textbf{56.0\%} 
 & \textbf{58.185\%}\\
  
  \bottomrule
  \end{tabular}
\end{table}

\begin{table}[!t]
  \caption{Comparison of pass@1 Metric Performance on the multilingual HumanEval-X (zero-shot, greedy-decoding)}  
  \centering
  \label{tab:humaneval_x}
  \begin{tabular}{lccccc}
  \toprule
   \textbf{Trained Model}    &  \textbf{Java}  &  \textbf{C++}  &  \textbf{JavaScript}  &  \textbf{Golang}  & \textbf{Average} \\
  \midrule
  CodeLlama-13B-Py-base & 43.3\% & 41.46\% & 34.76\% & 38.41\% & 29.27\%  \\
  \hline
  \sftcodecomp    &     50.0\%      &  39.02\%   &   47.56\%  &   40.23\%  & 44.20\%  \\
  \sftmix      &    56.1\%       &   \textbf{48.17\%}   &  \textbf{56.10\%}  &  37.80\% & 49.54\%  \\
  \rowcolor{gray!50}
  \mft          &     \textbf{57.32\%}    &   46.34\%   &  54.27\%   &  \textbf{45.12\%} & \textbf{50.76\%} \\
  \bottomrule
  \end{tabular}
\end{table}

\subsubsection{Text-to-Code Generation}

To evaluate the models' ability to generate code based on descriptions, we selected the MBPP~\cite{austin2021program} evaluation dataset and used the pass@1 metric. MBPP is specifically designed to assess models' capacity to synthesize concise Python programs from natural language descriptions.

We tested three models, namely \sfttextcode, \sftmix, and \mft, on the MBPP dataset, measuring their pass@1 performance as shown in Table~\ref{tab:humaneval_mbpp} (Column \textrm{IV}). Among these models, \mft exhibited the highest performance, surpassing the \sftmix model by 2.4\%. Similarly, in terms of the text-to-code generation task, models fine-tuned after combining multiple tasks showed inferior performance compared to models fine-tuned specifically for this individual task.

\textbf{Overall, in terms of text-to-code generation tasks, models trained using the MFT method outperform both
individually fine-tuned models and models fine-tuned after combining multiple tasks.}

\subsubsection{Code Comment Generation}

The objective of the code comment generation task is to have models add necessary comments to the code without modifying the input code itself. This includes both line comments and interface comments, making the code more readable and user-friendly. 

To assess this capability, we constructed an evaluation set based on 500 MBPP test questions (id xx-xx).
For each question in the evaluation set, we had the \sftcodecomment, \sftmix and \mft models generate comments for it. Subsequently, we employed GPT-4 as the referee, which has been instructed with criteria for good comments, to determine which model performed the best. If it was not possible to determine, the output was labeled as UNKNOWN. Finally, we counted the number of questions where each model was determined to perform the best and calculated the corresponding proportions, shown in Table~\ref{tab:codecomment}.

It can be observed that 38.8\% of the questions were determined to be best performed by the \mft model, surpassing the second-ranked \sftmix by 7.4\% and the third-ranked \sftcodecomment by 10.8\%. Additionally, 1.8\% of the questions were marked as indeterminable by GPT-4. 
\textbf{In summary, for this task, the models trained using the MFT method exhibit the best performance.}

\begin{table}[!t]
  \caption{Performance Comparison of Three Models on Code Commenting Task. GPT-4 Determines the Best Performing Model for Each Question. This Table Presents the Proportion of Questions Where Each Model Performs the Best. In particular, 1.8\% of the evaluation cases were indeterminate for GPT-4 to determine the best-performing model.}  
  \centering
  \label{tab:codecomment}
  \begin{tabular}{ccccccc}
  \toprule
   \textbf{Training Model}        &  \textbf{Best identified by GPT-4}  \\
  \midrule
  \sftcodecomment           &       28\%    \\
  \sftmix         &     31.4\%        \\
  \rowcolor{gray!50}
  \mft          &        \textbf{38.8\%}    \\
  \bottomrule
  \end{tabular}
\end{table}

\subsubsection{Code Translation}

The objective of the code translation task is to accurately and precisely translate a given code snippet implemented in the source language into an equivalent code snippet implemented in the target language while ensuring that both implementations possess identical functionality.
Here, we utilize the \textsc{codefuseEval}~\footnote{https://github.com/codefuse-ai/codefuse-evaluation}~\cite{di2023codefuse13b} evaluation datasets' code translation subset to support bidirectional translation between Java, Python, and C++. In order to evaluate the accuracy and functional equivalence of the translation results, we employ test cases that are semantically equivalent to ones of the source program for each task. These test cases are used to verify whether the result code can run and pass successfully, as indicated by the pass@1 criterion.

The test results of the three models are presented in Table~\ref{tab:codetrans}. The \mft model performs the best in Python-to-Java, Python-to-C++, and C++-to-Java translations. The \sftmix model excels in C++-to-Python translation, while the \sftcodetrans model performs the best in Java-to-Python and Java-to-C++ translations. Overall, the \mft model demonstrates superior performance, with an average improvement of 0.93\% over \sftmix and 10.9\% over \sftcodetrans. 
\textbf{This task also highlights the phenomenon that models trained using the MFT approach outperform the other two training methods.}

\begin{table*}[!t]
  \caption{Comparison of pass@1 Metric Performance on the \textsc{codefuseEVAL}-CodeTranslation
  ~\cite{di2023codefuse13b} (zero-shot, greedy-decoding)}  
  \centering
  \label{tab:codetrans}
  \begin{tabular}{lccccccc}
  \toprule
   \textbf{Training Model}        &  \textbf{Py2Java}  &  \textbf{Py2C++}  &  \textbf{Java2Py}  &  \textbf{C++2Py}  &  \textbf{Java2C++} & \textbf{C++2Java}  & \textbf{Avg.} \\
  \midrule
  \sftcodetrans           &       59.52\%       &     57.40\%      &  \textbf{70.73\%}   &   62.20\%  &   \textbf{67.07\%}  & 62.80\%  & 63.29\% \\
  \sftmix         &     80.16\%          &    71.20\%       &   67.68\%   &  \textbf{72.56\%}  &  65.85\% & 82.31\%  & 73.29\% \\
  \rowcolor{gray!50}
  \mft          &        \textbf{82.16\%}       &     \textbf{77.20\%}    &   65.85\%   &  70.73\%   &  64.64\% & \textbf{84.76\%}  & \textbf{74.22\%} \\
  \bottomrule
  \end{tabular}
\end{table*}

\subsubsection{Unit Test Case Generation}

The task at hand is to generate unit test cases by training a model to produce a set of test cases for a given code snippet, such as a method or class, and verify if the provided code implementation is correct. 
We have opted to utilize the unittest subset from the \textsc{codefuseEVAL}~\cite{di2023codefuse13b} evaluation datasets as our test suite. 
We evaluate the test cases using the pass@1 metric, which means that if a model generates test cases for a sample program and the sample program passes all the test cases, the count of correctly generated samples increases by 1. Similar to CodeLLama~\cite{roziere2023code}, we employ a greedy decoding strategy during the evaluation process.

We compared three models across Python, Java, and JavaScript for their test generation capabilities. The results in Table~\ref{tab:unittest} indicate that the \mft model outperforms others in test generation for Python, with a 5.73\% lead over the second-ranked \sftmix model, and a significant 10.19\% lead over the third-ranked \sftunit model.
In JavaScript, the \mft model also excels, with a 7.93\% advantage over the other models. However, in Java, the \mft model performs 5.37\% better than \sftunit, but slightly lags behind \sftmix by 5.44\%.
Overall, the \mft model consistently demonstrates the highest performance, with an average improvement of 2.74\% over \sftmix, and a remarkable 7.83\% improvement over \sftunit.



In summary, training models using the MFT approach yielded better performance compared to models fine-tuned with the same data mixed together. Furthermore, the performance of the MFT-trained models surpassed that of the model individually fine-tuned for the \textsc{UNIT-TEST} task.

\begin{table*}
  \caption{Comparison of pass@1 Metric Performance on the codefuseEVAL-TestcaseGeneration~\cite{di2023codefuse13b} (zero-shot, greedy-decoding)}  
  \centering
  \label{tab:unittest}
  \begin{tabular}{lcccc}
  \toprule
   \textbf{Trained Model}        &  \textbf{Python}  &  \textbf{Java}  &  \textbf{JavaScript}  & \textbf{Average} \\
  \midrule
  \sftunit           &       33.76\%       &     32.43\%      &  41.46\%   &   35.88\%    \\
  \sftmix        &     38.22\%          &    \textbf{43.24\%}       &   41.46\%   &  40.97\%    \\
  \hline
  \rowcolor{gray!50}
  \mft          &        \textbf{43.95\%}       &     37.8\%    &  \textbf{49.39\%}   &  \textbf{43.71\%}    \\
  \bottomrule
  \end{tabular}
\end{table*}

\subsubsection{Generalization on an Unseen Task}

In addition to evaluating the performance of our models on tasks with training data to address \textsc{RQ1} and \textsc{RQ2}, we also have \textsc{RQ3} to answer, which is whether the models trained using the MFT approach exhibit better generalization on unseen tasks compared to models trained by combining multiple datasets into a single SFT approach.
To investigate this, we selected the Text-to-SQL generation task as our testing target. The data for this task was not included in the training of the seven existing models. Furthermore, this task is code-related but distinctly different from the previous five downstream tasks. 

We have selected two evaluation metrics, BLEU score and logical accuracy of SQL statements. The BLEU score assesses the textual similarity between the generated outputs and the reference answers. The logical accuracy metric, on the other hand, allows us to address the variations in SQL syntax that may occur. Specifically, Logical Accuracy measures the proportion of test samples in the dataset where the generated SQL statements are both syntactically correct and semantically equivalent to the reference answers.

\begin{table}
  \caption{Sample Text-to-SQL Test Cases from the Spider~\cite{yu2019spider} dataset and Example Predictions by the SFT-Mixed Model and MFT-5Tasks Model.}

  \centering
  \label{tab:text_sql_example}
  \begin{tabular}{cl}
  \toprule
   \textcolor{blue}{PROMPT:} & \tabincell{p{0.8\textwidth}}{<|role\_start|>system<|role\_end|>You are a professional SQL engineer and you are writing SQL queries for data query tasks. \\ <|role\_start|>human<|role\_end|>What are the names of the heads who are born outside the California state? \\ schema:head(head\_ID|name|born\_state|age \\ <|role\_start|>bot<|role\_end|>}\\
   \hline
   \textcolor{green}{LABEL:} & \tabincell{p{0.8\textwidth}}{\ \\SELECT name FROM head WHERE born\_state != 'California'} \\
   \hline
   \sftmix: & \tabincell{p{0.8\textwidth}}{\ \\You can use the SQL SELECT statement to retrieve the names of the heads who are born outside the California state. Here is a sample query:\\ \\```sql\\SELECT name\\FROM head\\WHERE born\_state <> 'California';\\```\\ \\This query selects the name column from the head table where the born\_state column is not equal to 'California'.} \\
   \hline
   \mft: & \tabincell{p{0.8\textwidth}}{\\SELECT name \\FROM head \\WHERE born\_state <> 'California';} \\
  \bottomrule
  \end{tabular}
\end{table}

We selected five representative text-to-SQL datasets, including WikiSQL~\cite{zhongSeq2SQL2017}, Spider~\cite{yu2019spider}, CSpider~\cite{min2019pilot}, CoSQL~\cite{yu2019cosql}, and BirdSQL~\cite{li2023llm}, and randomly sampled 200 examples from each dataset for evaluation. The test case examples are shown in Table~\ref{tab:text_sql_example}, where the first row demonstrates the fine-tuned data format similar to OpenAI ChatML format~\footnote{https://github.com/openai/openai-python/blob/main/chatml.md}. Using each sampled dataset, we tested the logical accuracy and BLEU score of the \sftmix and \mft models, as shown in Table~\ref{tab:text_sql}.

According to Table~\ref{tab:text_sql},  \mft  outperforms \sftmix in terms of BLEU scores on each dataset, averaging 2.78 times higher. This indicates that the generated results of \mft exhibit higher similarity to the reference answer texts. This similarity can also be observed in Table~\ref{tab:text_sql_example}, where \mft produces cleaner results, while \sftmix provides more explanations, which may be preferred in certain scenarios. Moreover, \mft demonstrates better performance in terms of logical accuracy, achieving an overall accuracy that is 2.18 times higher than \sftmix model, and up to 4.67 times higher on the WikiSQL dataset. \textbf{Numerically, \mft exhibits superior performance compared to \sftmix, indicating stronger generalization of MFT-trained models on the Text-to-SQL task, which is an unseen task during training.}

\begin{table*}
  \caption{Comparison of generalization capabilities between \mft and \sftmix on the Text-to-SQL task. The evaluation metrics include SQL logical accuracy and BLEU score.}
  \centering
  \label{tab:text_sql}
  \begin{tabular}{lcccccc}
  \toprule
   \textbf{Trained Model}        &  \textbf{WIKISQL} &  \textbf{SPIDER} &  \textbf{CSPIDER}  &  \textbf{COSQL}  &  \textbf{BiRDSQL} & \textbf{Average} \\
  \midrule
  \multicolumn{7}{c}{Logical Accuracy} \\
  \hline
  \sftmix        &    1.5\%          &    2.0\%       &   7.0\%   &  6.5\%  &  5.5\% &  4.5\% \\
  \hline
  \rowcolor{gray!50}
  \mft          &        7.0\% (\textcolor{green}{4.67x})       &     4.5\% (\textcolor{green}{2.25x})    &   16.5\% (\textcolor{green}{2.36x})   &  10.5\%(\textcolor{green}{1.62x})   &  10.5\% (\textcolor{green}{1.91x}) &  9.8\% (\textcolor{green}{2.18x}) \\
  \midrule
  \multicolumn{7}{c}{BLEU} \\
  \hline
  \sftmix & 0.032 & 0.047 & 0.025 & 0.081 & 0.026 & 0.042 \\

  \rowcolor{gray!50}
  \mft & 0.138 & 0.138 & 0.116 & 0.119 & 0.074 & 0.117\\
 
  \bottomrule
  \end{tabular}
\end{table*}

\subsection{Evaluation Summary}

We selected five downstream tasks related to code and trained a total of seven models, including \textsc{SFT-S-*} models fine-tuned individually for each task, the \sftmix model fine-tuned with a mixture of all task data, and the \mft model trained using the MFT method. We compared and tested the performance of each model in terms of their target capabilities. Additionally, we evaluated the generalization performance of the MFT method and the mixed SFT method on unseen tasks. The results can be summarized as follows:

\begin{enumerate}[label=\roman*, leftmargin=*]

 \item \textbf{Models trained with the MFT method outperformed those fine-tuned individually for each task, indicating a positive answer to RQ1.}

\item \textbf{Models trained with the MFT method outperformed those fine-tuned with a mixture of multiple tasks, providing a positive answer to RQ2.}

\item \textbf{Models trained with the MFT method exhibit stronger generalization capabilities on new, unseen tasks compared to the SFT models fine-tuned with a mixture of multiple task data.}

\end{enumerate}

\section{Application}

Considering the outstanding performance of the MFT training method, we have leveraged our \sysname~\textsuperscript{\ref{mftcoder}}, developed based on this approach, to fine-tune the existing mainstream open-source LLM models. \eg QWen~\cite{qwen}, Baichuan~\cite{baichuan2023baichuan2}, CodeGeex2~\cite{zheng2023codegeex}, Llama~\cite{touvron2023llama}, LLama2~\cite{touvron2023llama2}, CodeLLama~\cite{roziere2023code}, StarCoder~\cite{li2023starcoder}.

\sysname supports Lora and QLora, which significantly reduces the number of model training parameters. Coupled with dual quantization for model size compression, this ultimately leads to a substantial reduction in GPU memory requirements. As a result, it becomes possible to fine-tune a 70B model on a single A100 GPU with ease.
When fine-tuning these models using \sysname, we set the trainable parameters to be within the range of 0.1\% to 5\% of the total parameters. Through empirical evidence, we have found that as the proportion of trainable parameters increases, performance improvement tends to plateau. In fact, we have observed that a trainable parameter proportion of less than 5\% is often sufficient to achieve performance levels close to that of full-scale fine-tuning.

When fine-tuning these models, we configure them for multitasking with a range of 3-7 tasks. Depending on the model size, we typically use Lora mode for models below 20B, and QLora mode for models larger than 20B. After fine-tuning, we evaluate their performance in code completion and text-to-code generation tasks, measuring their performance on HumanEval~\cite{chen2021codex} and MBPP~\cite{austin2021program}, as shown in Table~\ref{tab:app_humaneval_mbpp} Column~\textrm{III} and \textrm{IV}.
We have calculated the average improvement of MFT fine-tuning compared to the base models in terms of HumanEval and MBPP. As shown in column 5, the improvement ranges from 6.26\% to 12.75\%, with the improvements on HumanEval consistently surpassing those on MBPP. 
Additionally, we have also evaluated the code completion performance of the \sysname fine-tuned models on the multilingual benchmark, HumanEval-X~\cite{zheng2023codegeex}. The results of this evaluation are presented in Table~\ref{tab:app_humaneval_x}.
Notably, the fine-tuned CodeFuse-CodeLLama-Python-MFT (34B) achieved an average pass@1 of 56.88\% across four languages: Java, C++, JavaScript, and Golang.

\begin{figure}[!t]
    \centering
    \includegraphics[width=\textwidth]{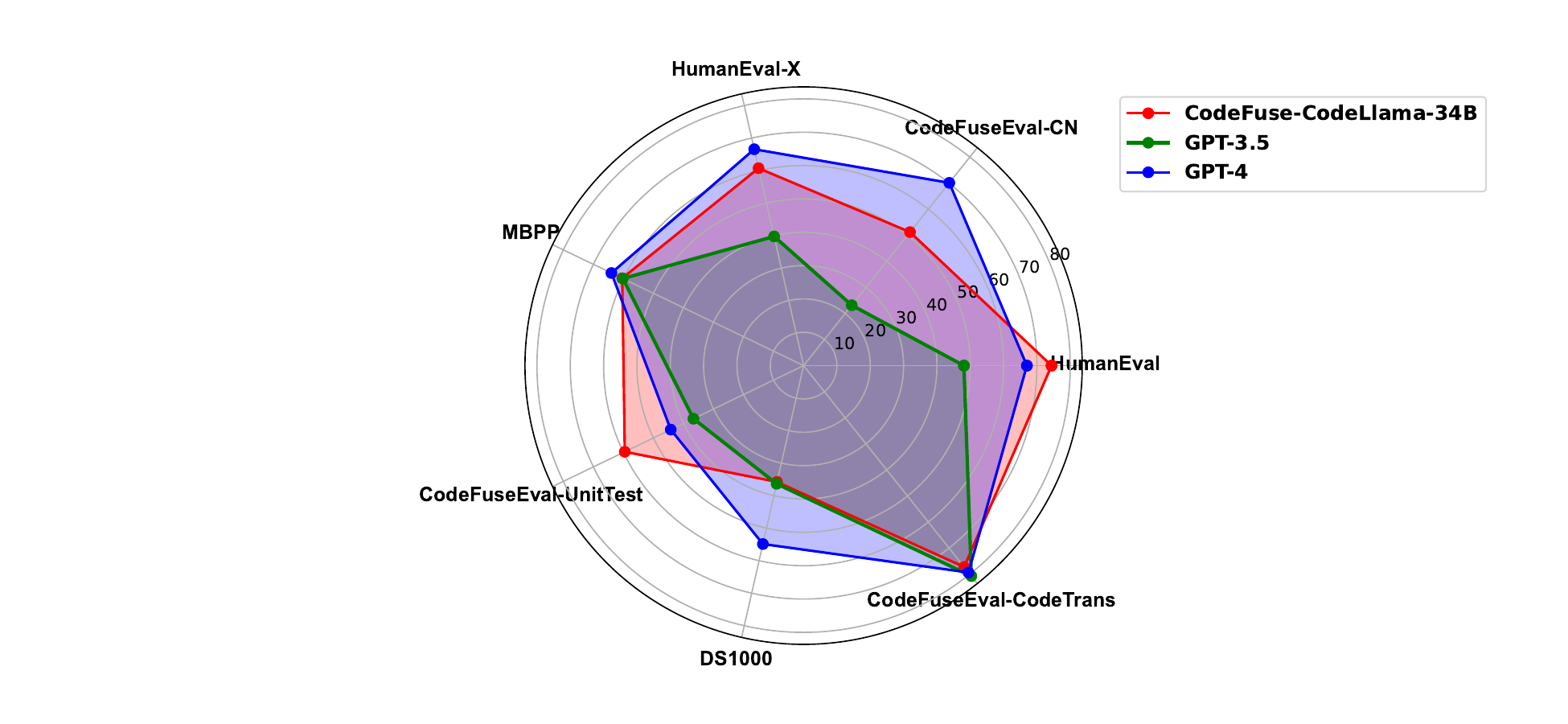}
    \caption{
     Radar Chart of CodeFuse-CodeLlama-34B Model on \textsc{HumanEval}, \textsc{HumanEval-X}, \textsc{MBPP}, \textsc{DS-1000}, and \textsc{codefuseEval} benchmarks compared to GPT-3.5 and GPT-4.}
    \label{fig:codefuse_codellama_34b_abilities}
\end{figure}

Table~\ref{tab:app_humaneval_mbpp} also presents the performance of fine-tuned open-source models (\eg OctoPack~\cite{muennighoff2023octopack} and WizardCoder-Python~\cite{luo2023wizardcoder}) and representative closed-source models (e.g., Claude2~\cite{anthropic2023claude}, GPT-4~\cite{openai2023gpt4}) on HumanEval and MBPP. \textbf{It is worth noting that our fine-tuned model, CodeFuse-CodeLLama-34B~\footnote{\label{codefuse_codellama_34b}https://huggingface.co/codefuse-ai/CodeFuse-CodeLlama-34B}, based on CodeLlama-34B-Python achieves a remarkable performance of 74.4\% on HumanEval, surpassing all the listed models in the table, including GPT-4 (67.00\%, zero-shot)~\cite{openai2023gpt4}.}
We also evaluated the performance of the model on other benchmarks, including multilingual \textsc{HumanEval-X}~\cite{zheng2023codegeex}, \textsc{MBPP}~\cite{austin2021program}, \textsc{DS-1000}~\cite{Lai2022DS1000}and \textsc{codefuseEval}~\cite{di2023codefuse13b}, and compared it against GPT-3.5 and GPT-4, as shown in Figure~\ref{fig:codefuse_codellama_34b_abilities}. 
CodeFuse-CodeLLama-34B outperforms GPT-4 on \textsc{CodeFuseEval-UnitTest} and \textsc{Humaneval}, matches its performance in code translation ability, but falls behind in Chinese code completion (\textsc{CodeFuseEval-CN}), multi-language completion, data-science analysis (\textsc{DS-1000}), and text-to-code generation (\textsc{MBPP}) capabilities compared to GPT-4. However, it surpasses or equals GPT-3.5 on all evaluation datasets.
The input-output examples on each evaluation dataset can be found in Appendix~\ref{apx:becnhamrk_example}.

\begin{figure}[!t]
    \centering
    \includegraphics[width=0.6\textwidth]{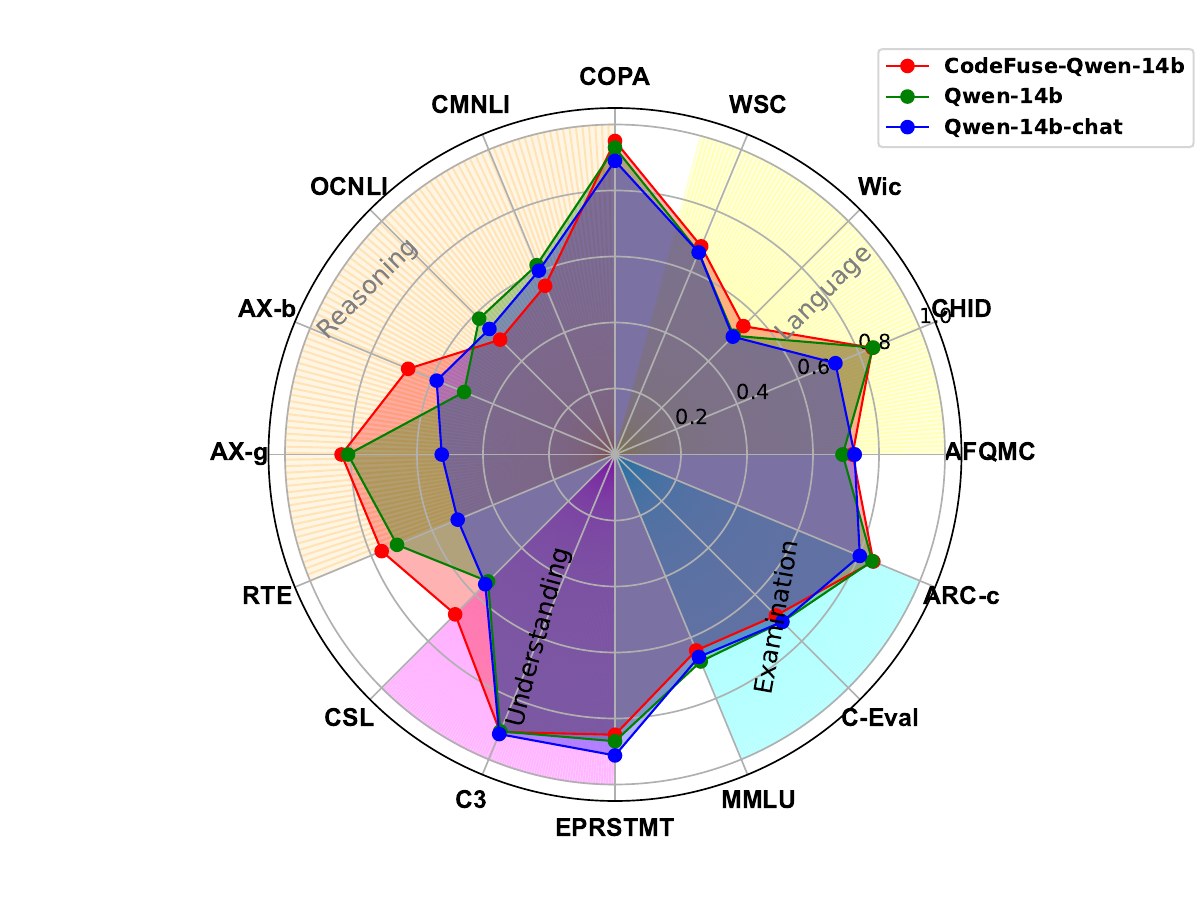}
    \caption{
     Performance comparison of \textsc{CodeFuse-QWen-14B} fine-tuned with \sysname and code-related data, \textsc{QWen-14B} base model, and officially fine-tuned model \textsc{QWen-14B-chat} on NLP evaluation datasets. Detailed data can be found in Appendix~\ref{apx:codefuse_qwen_nlp_data}.}
    \label{fig:qwen_nlp_radar}
\end{figure}

Furthermore, we conducted an evaluation to assess the impact of fine-tuning the models with \sysname and code-related data on their performance in NLP tasks, as illustrated in Figure~\ref{fig:qwen_nlp_radar}. Taking \textsc{CodeFuse-QWen-14B} as a case study, we compared it against the base model \textsc{QWen-14B} and the official model \textsc{QWen-14B-chat} fine-tuned by Alibaba Cloud on top of it. It is evident that \textsc{CodeFuse-QWen-14B} maintains its proficiency in NLP. In fact, it exhibits a slight enhancement in language, reasoning, and understanding abilities compared to the other two models. However, there is a minor decline in its examination ability when compared to the base model \textsc{QWen-14B} and similar findings are observed for the fine-tuned \textsc{QWen-14B-chat} model.

\begin{table}[!t]
  \caption{pass@1 performance on HumanEval~\cite{chen2021codex} (Code Completion) and MBPP~\cite{austin2021program} (Text-to-Code Generation) after fine-tuning with \sysname across multiple mainstream open-source models. The \textit{CodeFuse-*-MFT} models are evaluated using a combination of greedy decoding and zero-shot testing strategy, while the metric values for the other models are taken from their respective papers, technical reports, or open-source project homepages.} 
  \centering
  \label{tab:app_humaneval_mbpp}
  \begin{tabular}{lcccl}
  \toprule
   \textbf{Model} &  \textbf{Size} &  \textbf{\tabincell{c} {Humaneval \\ pass@1}}  & \textbf{\tabincell{c}{MBPP \\ pass@1}} & \textbf{Average} \\
  \midrule
  \multicolumn{4}{c}{Open-source base models} \\
  \hline
  
  QWen-base~\cite{qwen}  & 14B & 32.3\%  & 40.8\% &  36.55\% \\
 Llama-base~\cite{touvron2023llama} & 65B & 23.7\% & 37.7\%  & 30.7\% \\
 Llama2-base~\cite{touvron2023llama2} & 70B & 29.9\% & 45.0\% & 37.45\% \\
 StarCoder-base~\cite{li2023starcoder} & 15B & 33.6\% & 52.7\% & 43.15\% \\
 CodeGeex2-base~\cite{zheng2023codegeex} & 6B &  35.9\% & 42.4\% & 39.15\% \\
 CodeLlama-Python-base~\cite{roziere2023code} & 13B & 43.3\% & 49.0\% & 46.15\% \\
 CodeLlama-Python-base~\cite{roziere2023code} & 34B &  53.7\% & 56.2\% & 54.95\% \\
  \hline
  \multicolumn{4}{c}{MFT fine-tuned models} \\
  \hline
  \rowcolor{gray!10}
  CodeFuse-QWen-MFT~\tablefootnote{https://huggingface.co/codefuse-ai/CodeFuse-QWen-14B} & 14B & 48.78\% &  43.8\% & 46.29\% (\textcolor{green}{+9.74\%}) \\
\rowcolor{gray!10}
 CodeFuse-Llama-MFT & 65B &  34.76\% & 41.8\% & 38.28\% (\textcolor{green}{+7.58}) \\
\rowcolor{gray!10}
 CodeFuse-Llama2-MFT & 70B & 40.85\% & 40.8\% & 40.83\% (\textcolor{green}{+3.38\%})\\
\rowcolor{gray!10}
 CodeFuse-StarCoder-MFT~\tablefootnote{https://huggingface.co/codefuse-ai/CodeFuse-StarCoder-15B} & 15B & 54.90\% &  49.60\% & 52.25\% (\textcolor{green}{+9.10\%}) \\
\rowcolor{gray!10}
 CodeFuse-CodeGeex2-MFT & 6B & 45.12\%  & 46.2\% & 45.66\% (\textcolor{green}{+6.51\%}) \\
\rowcolor{gray!10}
 CodeFuse-CodeLlama-Python-MFT & 13B & 60.37\% &  56.0\%  & 58.19\% (\textcolor{green}{+12.04\%})\\
\rowcolor{gray!10}
 CodeFuse-CodeLLama-Python-MFT~\textsuperscript{\ref{codefuse_codellama_34b}} & 34B & \textbf{74.4\%} & \textbf{61.0\%} & \textbf{67.70\%} (\textcolor{green}{+12.75\%})\\

 \hline
 \multicolumn{4}{c}{Open-source fine-tuned models} \\
 \hline
 QWen-chat~\cite{qwen} & 14B &  43.9\% & 46.4\% & 45.15\%  \\
 
  PHI-1~\cite{gunasekar2023textbooks} & 1.3B & 50.6\% &  55.5\% & 53.05\% \\ 
  OctoCoder~\cite{muennighoff2023octopack} & 15B & 46.2\% & NA & NA \\
  WizardCoder~\cite{luo2023wizardcoder} & 15B & 57.3\% &  51.8\% & 54.55\% \\

  Phind-CodeLlama-v2~\cite{phind2023codellama} & 34B & 71.95\% & NA & NA \\
  WizardCoder-Python~\cite{luo2023wizardcoder} & 34B & 73.2\% & 61.2\% & 67.2\% \\
  \hline
  \multicolumn{4}{c}{Closed-source models} \\
  \hline
  
  PanGu-Coder2~\cite{shen2023pangu} & 15B & 61.2\% & 
 NA & NA \\
  Unnatural CodeLlama~\cite{roziere2023code} & 34B & 62.2\% & 61.2\% & 61.7\% \\
  Claude2~\cite{anthropic2023claude} & NA & 71.2\% & NA & NA \\
  GPT-3.5~\cite{openai2023gpt4} & 175B & 48.1\%  & 52.2\% & 50.15\% \\
  GPT-4 (zero-shot)~\cite{openai2023gpt4} & NA & 67.00\% & NA & NA \\
  
  \bottomrule
  \end{tabular}
\end{table}

\begin{table}[!t]
  \caption{pass@1 performance on Multi-lingual HumanEval-X~\citep{zheng2023codegeex} after fine-tuning with \sysname across multiple mainstream open-source models. The metric values marked with an asterisk (*) were obtained from the models' corresponding papers, technical reports, or open-source project homepages, while the remaining metric values were evaluated using a combination of greedy decoding and zero-shot testing strategy.} 
  \centering
  \label{tab:app_humaneval_x}
  \begin{tabular}{lcccccccc}
  \toprule
   \textbf{Model}   & \textbf{Size}  &  \textbf{Python} & \textbf{Java}  &  \textbf{C++}  &  \textbf{JavaScript}  &  \textbf{Golang}  & \textbf{Avgerage} \\
  \midrule
  QWen-base  &14B & $32.3\%^{*}$  &  35.37\% & 30.49\% & 32.93\% & 21.34\% & 30.49\% \\
  CodeFuse-QWen-MFT & 14B  & 48.78\% & 41.46\% & 38.41\% & 46.34\%
 & 26.83\% & 40.36\% \\
 \hline
 \midrule
 Llama-base & 65B & $23.7\%^{*}$ &  29.26\% & 20.73\% & 23.78\% & 18.9\% & 23.27\%  \\
 CodeFuse-Llama-MFT & 65B  & 34.76\% & 37.2\% &  29.88\% & 32.93\% & 23.78\% & 31.71\% \\
 \hline
 \midrule
 Llama2-base & 70B & $29.9\%^{*}$  & 39.02\% & 31.10\% & 
 35.98\% &  23.78\% & 31.96\% \\
 CodeFuse-Llama2-MFT & 70B & 40.85\% & 35.98\% & 32.32\% & 38.41\% & 27.44\% & 35.00\% \\
 \hline
 \midrule
 StarCoder-base & 15B & $33.6\%^{*}$  & 34.15\% & 25.61\% & 22.56\% & 22.56\% & 29.48\% \\
 CodeFuse-StarCoder-MFT & 15B &  54.9\%  & 47.56 & 46.34\% & 48.17\% & 37.20\% & 46.83\% \\
 \hline
 \midrule
 CodeGeex2-base & 6B &  $35.9\%^*$  &  $30.8\%^*$ & $29.3\%^*$ & $32.2\%^*$ & $22.5\%^*$ & 30.14\% \\
 CodeFuse-CodeGeex2-MFT & 6B &  45.12\%  & 45.73\% & 37.2\%  & 37.2\% & 28.05\% & 38.66\% \\
 \hline
 \midrule
 CodeLlama-Python-base & 13B  & $43.3\%^*$ & 41.46\% & 34.76\% & 38.41\% & 29.27\% & 37.44\% \\
 CodeFuse-CodeLlama-Python-MFT & 13B & 60.37\%  & 57.32\% & 46.34\% & 54.27\%	& 45.12\% & 52.68\% \\
 \hline
 \midrule
 CodeLlama-34B-Python-base & 34B & $53.7\%^*$ &  45.73\%	&  42.68\% & 45.73\% & 31.71\% & 43.91\%  \\
 CodeFuse-CodeLLama-Python-MFT & 34B & 	\textbf{74.4\%} &\textbf{61.6\%}  &	\textbf{54.3\%}  & 	\textbf{61.0\%} 	& \textbf{50.6\%}  & \textbf{60.38\%}\\

  \bottomrule
  \end{tabular}
\end{table}

\section{Discussion}

Despite the superior performance of the MFT training method compared to the task data mixing-based SFT training method in the aforementioned experiments, it should be noted that the effectiveness of the MFT approach is highly dependent on the task-splitting strategy. Not all scenarios are suitable for being split into multiple tasks. For instance, in our practical experience, we found that splitting a task based on difficulty levels and training it using the MFT method did not yield better results compared to the task-mixed SFT training method. Moreover, training code completion tasks as multiple tasks based on programming languages also did not outperform the mixed SFT method. Based on our practical experience, we have concluded that tasks with distinct main desired abilities are more suitable for task splitting and MFT training, whereas tasks with similar main training objectives are not well-suited for MFT training. We plan to further investigate and establish more precise criteria for task delineation in future research.

In our task generalization experiments, we observed that models trained using the MFT method produced inference results that were more similar to the reference answers and had more concise content. Conversely, inference results generated by the task-mixed SFT training method contained more Chain-of-Thought (CoT) information. In certain scenarios, the former approach was more preferred, such as in IDE plugins, while the latter approach was favored in other scenarios, such as web assistants. As a result, we cannot simply generalize that one method is better than the other. {We are currently researching the reasons behind these performance differences.}

{As a multi-task learning method, MFT also faces a major challenge during the training process: inconsistent convergence speeds among different tasks. For example, in the aforementioned experiments, the code completion task converged much faster than the unit test-case generation task (details can be found in Appendix~\ref{apx:tasks_loss}). This makes it difficult to find an optimal point that performs well on all tasks. The selected checkpoint either converges insufficiently on some tasks or overfits on others. To address this issue, we experimented with existing multi-task learning balancing optimization solutions such as FAMO~\cite{liu2023famo}. However, FAMO requires dual back-propagation in each iteration, resulting in training time being approximately doubled. Furthermore, the required number of epochs for convergence also increases significantly, and the adjustability of the convergence speed is limited.
Unfortunately, this exponentially increased cost does not yield equivalent benefits. In response, we are currently developing a more optimal and adaptive multi-task optimization balancing approach.}

Furthermore, even after balancing the convergence speeds of multiple tasks, where the same set of parameters is updated, it is still challenging to fundamentally eliminate the inherent conflicts in weight updates across different tasks. To address this issue, we are currently exploring the utilization of MoE (Mixture of Experts)~\cite{chen2022understanding} to achieve MFT.

\section{Conclusion}

This paper introduces \sysname, a framework that supports multi-task fine-tuning, effectively addressing the challenges of data imbalance, varying difficulty levels, and inconsistent convergence speeds through the design of various loss functions. Experimental results demonstrate that this approach outperforms individual fine-tuning on each task or fine-tuning on a mixed ensemble of tasks. Additionally, \sysname facilitates efficient training, including efficient data utilization and PEFT training. It also provides a high-quality instruction dataset construction solution.
Leveraging \sysname for fine-tuning on the CodeLLama base, the CodeFuse-CodeLLama-34B model achieves an impressive pass@1 score of 74.4\% on the HumanEval dataset, surpassing the performance of GPT-4 (67\%, zero-shot).

\clearpage

\bibliographystyle{ACM-Reference-Format}
\bibliography{ref}
\clearpage

\appendix

\section{Code Exercises Generation with Camel}
\label{apx:camel}

\begin{table}[ht]
  \caption{Examples of Prompts for Code Exercises Generation using Camel}
  \scriptsize
  \centering
  \label{tab:camel_prompt}
  {\fontsize{7}{8}\selectfont
  \begin{tabular}{|c|c|}
  \toprule
  \multicolumn{2}{|c|}{\tabincell{p{\textwidth}}{Original task prompt:\\\\
  Create 50 coding exercises to test and improve students' Python programming skills.  The exercises must must must focus on {Binary search}. The difficulty of exercises must be from easy to medium to hard.}} \\
  \hline

  \multicolumn{2}{|c|}{\tabincell{p{\textwidth}}{Specified task prompt:\\\\
  A student will collaborate with a programming teacher to develop 50 Python coding exercises centered specifically on Binary search algorithms. The exercises should progressively increase in difficulty, from easy to medium to hard, targeting various aspects of implementation and optimization techniques.}} \\
  \hline
  
  \multicolumn{2}{|c|}{\tabincell{p{\textwidth}}{Final task prompt:\\\\
  A student will collaborate with a programming teacher to develop 50 Python coding exercises centered specifically on Binary search algorithms. The exercises should progressively increase in difficulty, from easy to medium to hard, targeting various aspects of implementation and optimization techniques.}} \\
  \hline
  
  \tabincell{p{0.5\textwidth}}{AI Assistant sys message:\\\\ SystemMessage(role\_name='A student', role\_type = <RoleType.ASSISTANT: 'assistant'>, meta\_dict=\{'task': 'A student will assist a programming teacher in designing a series of 50 Python coding exercises that emphasize the application and optimization of Bubble Sort algorithm. These exercises will span varying levels of difficulty, from easy to medium to hard, allowing students to strengthen their sorting skills progressively.', 'assistant\_role': 'A student', 'user\_role': 'A programming teacher'\}, role='system', content='Never forget you are a A student and I am a A programming teacher. Never flip roles! Never instruct me!\\We share a common interest in collaborating to successfully complete a task.\\You must help me to complete the task.\\Here is the task: A student will assist a programming teacher in designing a series of 50 Python coding exercises that emphasize the application and optimization of Bubble Sort algorithm. These exercises will span varying levels of difficulty, from easy to medium to hard, allowing students to strengthen their sorting skills progressively.. Never forget our task!\\I must instruct you based on your expertise and my needs to complete the task.\\\\I must give you one instruction at a time.\\You must write a specific solution that appropriately solves the requested instruction and explain your solutions.\\You must decline my instruction honestly if you cannot perform the instruction due to physical, moral, legal reasons or your capability and explain the reasons.\\Unless I say the task is completed, you should always start with:\\\\Solution: <YOUR\_SOLUTION>\\\\<YOUR\_SOLUTION> should be very specific, include detailed explanations and provide preferable detailed implementations and examples and lists for task-solving.\\Always end <YOUR\_SOLUTION> with: Next request.')\\\\\\\\\\\\\\\\\\\\\\\\\\\\\\} & 
  \tabincell{p{0.5\textwidth}}{AI User sys message:\\\\
  SystemMessage(role\_name='A programming teacher', role\_type = <RoleType.USER: 'user'>, meta\_dict = \{'task': 'A student will assist a programming teacher in designing a series of 50 Python coding exercises that emphasize the application and optimization of Bubble Sort algorithm. These exercises will span varying levels of difficulty, from easy to medium to hard, allowing students to strengthen their sorting skills progressively.', 'assistant\_role': 'A student', 'user\_role': 'A programming teacher'\}, role='system', content='Never forget you are a A programming teacher and I am a A student. Never flip roles! You will always instruct me.\\We share a common interest in collaborating to successfully complete a task.\\I must help you to complete the task.\\Here is the task: A student will assist a programming teacher in designing a series of 50 Python coding exercises that emphasize the application and optimization of Bubble Sort algorithm. These exercises will span varying levels of difficulty, from easy to medium to hard, allowing students to strengthen their sorting skills progressively.. Never forget our task!\\You must instruct me based on my expertise and your needs to solve the task ONLY in the following two 
  ways:\\\\1. Instruct with a necessary input:\\Instruction: <YOUR\_INSTRUCTION>\\Input: <YOUR\_INPUT>\\\\2. Instruct without any input:\\Instruction: <YOUR\_INSTRUCTION>\\Input: None\\\\The "Instruction" describes a task or question. The paired "Input" provides further context or information for the requested "Instruction".\\\\You must give me one instruction at a time.\\I must write a response that appropriately solves the requested instruction.\\I must decline your instruction honestly if I cannot perform the instruction due to physical, moral, legal reasons or my capability and explain the reasons.\\You should instruct me not ask me questions.\\Now you must start to instruct me using the two ways described above.\\Do not add anything else other than your instruction and the optional corresponding input!\\Keep giving me instructions and necessary inputs until you think the task is completed.\\When the task is completed, you must only reply with a single word <CAMEL\_TASK\_DONE>.\\Never say <CAMEL\_TASK\_DONE> unless my responses have solved your task.')\\ \\ \\ \\}\\

  \bottomrule
  \end{tabular}}
\end{table}
\clearpage

\section{Convergence Speed of MFT Training Process}
\label{apx:tasks_loss}

\begin{figure}[ht]
    \centering
    \begin{subfigure}{0.43\textwidth}
        \centering
        \includegraphics[width=\linewidth]{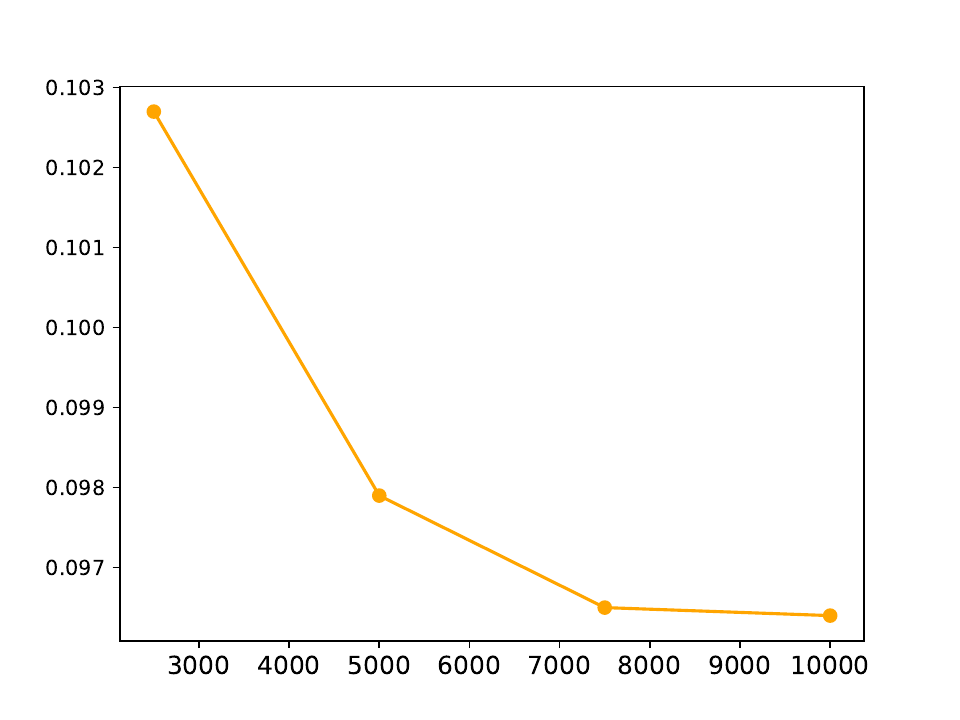}
        \caption{Code Comment Task}
        \label{fig:codecomment_loss}
    \end{subfigure}
    \hfill
    \begin{subfigure}{0.43\textwidth}
        \centering
        \includegraphics[width=\linewidth]{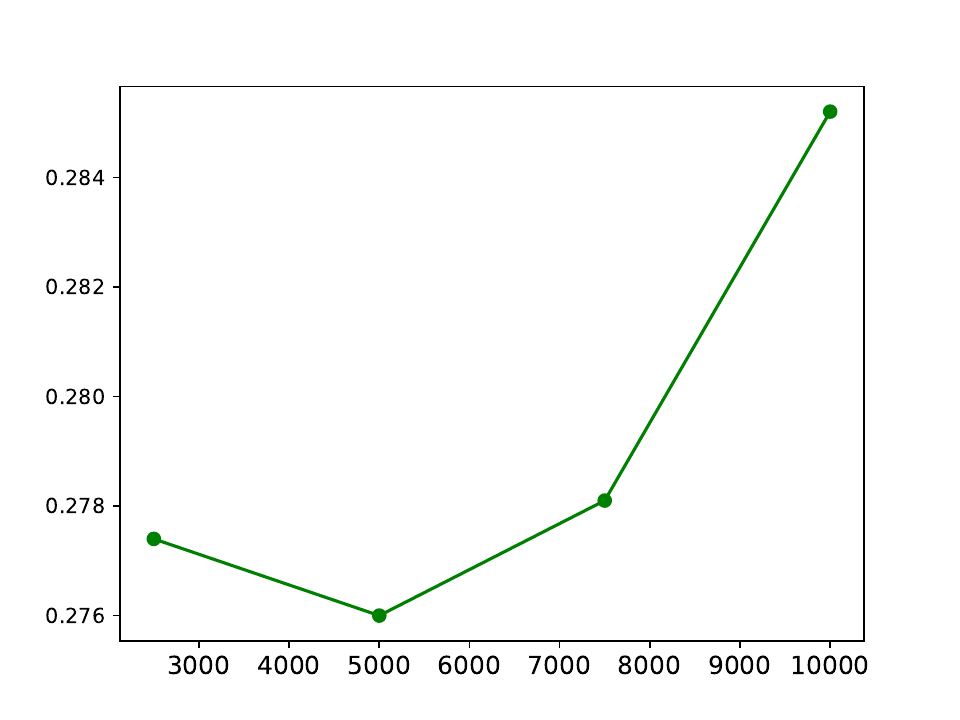}
        \caption{Code Completion Task}
        \label{fig:codecompl_loss}
    \end{subfigure}
    \hfill
    \begin{subfigure}{0.43\textwidth}
        \centering
        \includegraphics[width=\linewidth]{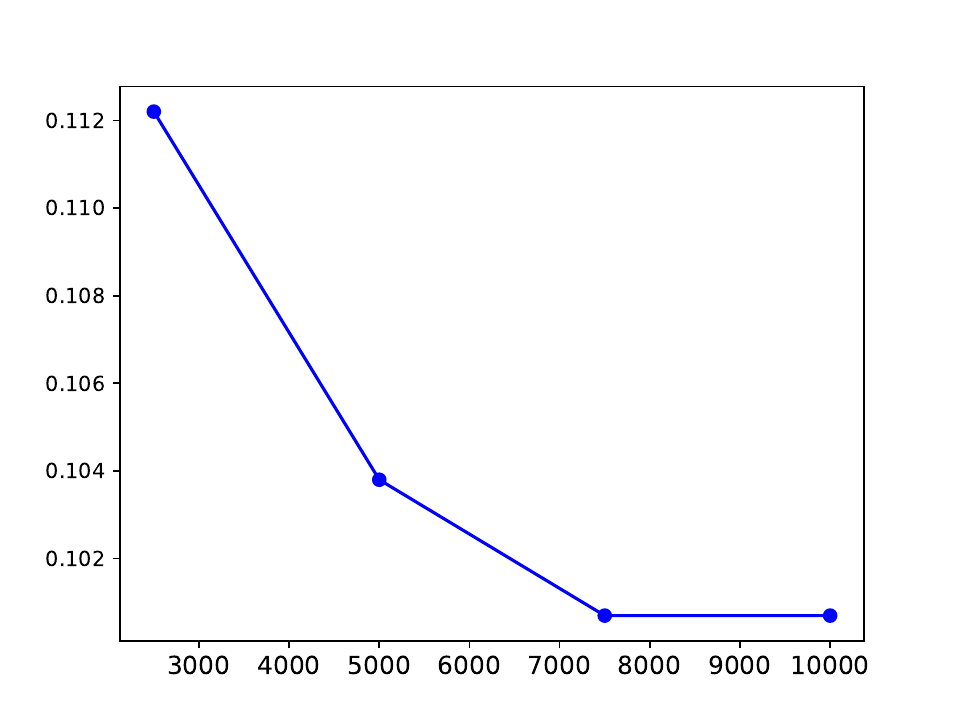}
        \caption{Code Translation Task}
        \label{fig:codetrans_loss}
    \end{subfigure}
    \hfill
    \begin{subfigure}{0.43\textwidth}
        \centering
        \includegraphics[width=\linewidth]{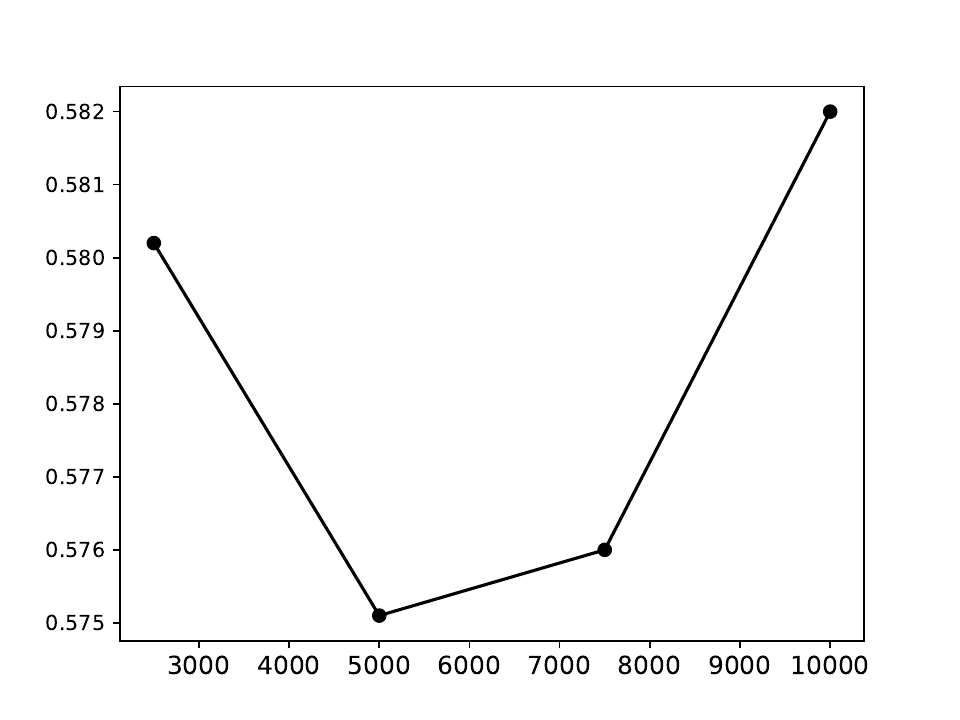}
        \caption{Text-to-Code Task}
        \label{fig:text2code_loss}
    \end{subfigure}
    \hfill
    \begin{subfigure}{0.43\textwidth}
        \centering
        \includegraphics[width=\linewidth]{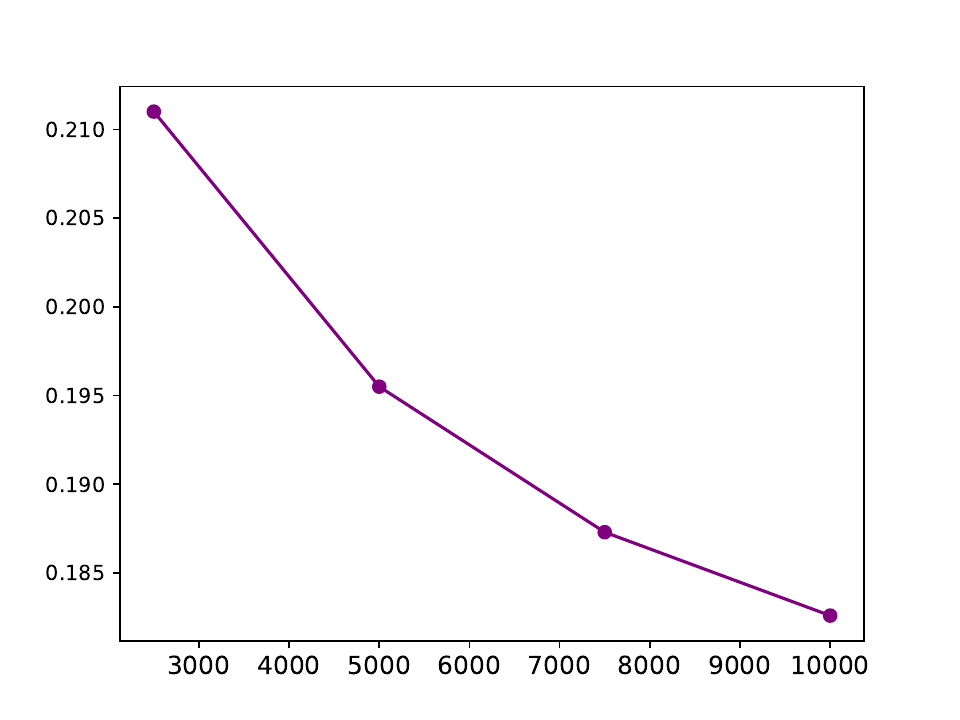}
        \caption{Unit Test-case Generation Task}
        \label{fig:unittest_loss}
    \end{subfigure}
    \hfill
    \begin{subfigure}{0.43\textwidth}
        \centering
        \includegraphics[width=\linewidth]{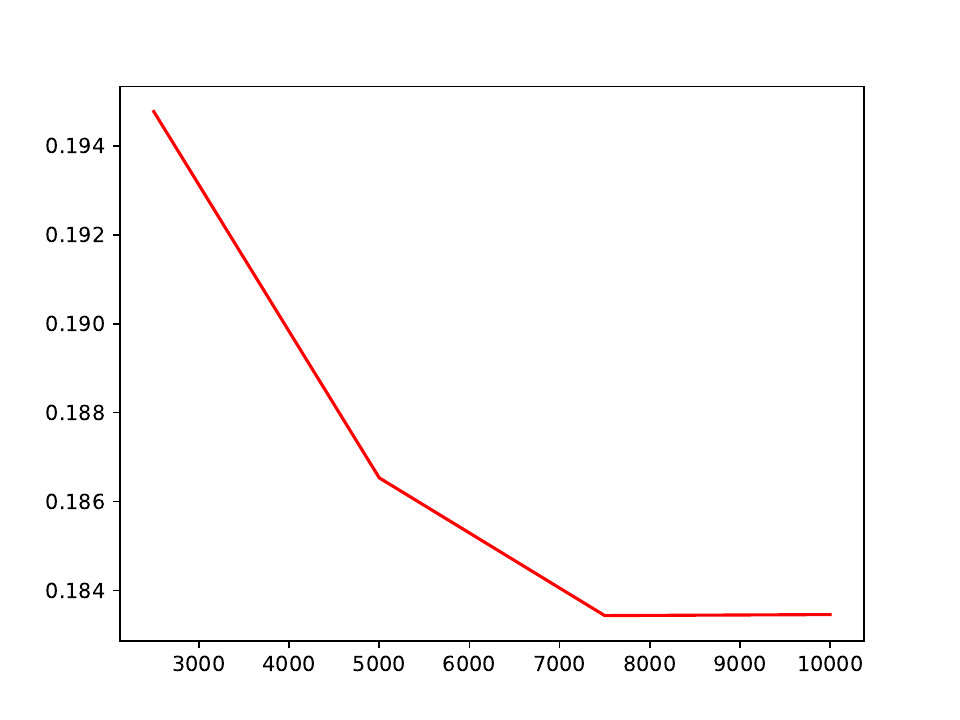}
        \caption{Overall of 5 tasks}
        \label{fig:overall_loss}
    \end{subfigure}
    \caption{Validation Loss Convergence Speeds: A Comparative Analysis of 5 Code-related Downstream Tasks and Overall Training Progress using the data-balanced loss function.}
    \label{fig:val_loss_codecompl_vs_unittest}
\end{figure}

\section{Examples of codefuse-codellama-34b across various tasks}
\label{apx:becnhamrk_example}

\subsection{Code Completion on HumanEval}

\lstset{keywordstyle=\color{blue}, commentstyle=\color{purple}, stringstyle=\color{red}}
\lstset{tabsize=4, aboveskip=1em, belowskip=1em}
\lstset{basicstyle=\ttfamily\small}

\begin{tcolorbox}[colback=yellow!5!white,colframe=yellow!50!black,
  colbacktitle=yellow!75!black,title=Prompt:]
<|role\_start|>human<|role\_end|>\# language: Python
\begin{lstlisting}[language=Python, breaklines=true]
from typing import List

def below_zero(operations: List[int]) -> bool:
    """ You're given a list of deposit and withdrawal operations on a bank account that starts with
    zero balance. Your task is to detect if at any point the balance of account fallls below zero, and
    at that point function should return True. Otherwise it should return False.
    >>> below_zero([1, 2, 3])
    False
    >>> below_zero([1, 2, -4, 5])
    True
    """
\end{lstlisting}
<|role\_start|>bot<|role\_end|>
\end{tcolorbox}

\begin{tcolorbox}[enhanced,frame style image=blueshade.png,
  opacityback=0.75,opacitybacktitle=0.25,
  colback=orange!5!white,colframe=orange!75!black,
  title=CodeFuse-CodeLLama-34B:]
\begin{lstlisting}[language=Python, breaklines=true]
    balance = 0
    for op in operations:
        balance += op
        if balance < 0:
            return True
    return False      
\end{lstlisting}
\end{tcolorbox}

\subsection{Multilingual Code Completion on HumanEval-X}

\begin{tcolorbox}
[colback=yellow!5!white,colframe=yellow!50!black,
  colbacktitle=yellow!75!black,title=Prompt (Java):]
<|role\_start|>human<|role\_end|>// language: Java
\begin{lstlisting}[language=Java, breaklines=true]
import java.util.*;
import java.lang.*;

class Solution {
    /**
    Given a string s, count the number of uppercase vowels in even indices.
    
    For example:
    countUpper("aBCdEf") returns 1
    countUpper("abcdefg") returns 0
    countUpper("dBBE") returns 0
     */    
    public int countUpper(String s) {
\end{lstlisting}
<|role\_start|>bot<|role\_end|>
\end{tcolorbox}

\begin{tcolorbox}[enhanced,frame style image=blueshade.png,
  opacityback=0.75,opacitybacktitle=0.25,
  colback=orange!5!white,colframe=orange!75!black,
  title=CodeFuse-CodeLLama-34B:]
\begin{lstlisting}[language=Java, breaklines=true]
    int count = 0;
        for (int i = 0; i < s.length(); i += 2) {
            char c = s.charAt(i);
            if (Character.isUpperCase(c) && (c == 'A' || c == 'E' || c == 'I' || c == 'O' || c == 'U')) {
                count++;
            }
        }
        return count;
    }
}      
\end{lstlisting}
\end{tcolorbox}

\begin{tcolorbox}
[colback=yellow!5!white,colframe=yellow!50!black,
  colbacktitle=yellow!75!black,title=Prompt (Golang):]
<|role\_start|>human<|role\_end|>// language: Go
\begin{lstlisting}[language=Go, breaklines=true]
import (
    "strings"
)

// Create a function Encrypt that takes a string as an argument and
// returns a string Encrypted with the alphabet being rotated.
// The alphabet should be rotated in a manner such that the letters
// shift down by two multiplied to two places.
// For example:
// Encrypt('hi') returns 'lm'
// Encrypt('asdfghjkl') returns 'ewhjklnop'
// Encrypt('gf') returns 'kj'
// Encrypt('et') returns 'ix'
func Encrypt(s string) string {
\end{lstlisting}
<|role\_start|>bot<|role\_end|>
\end{tcolorbox}

\begin{tcolorbox}[enhanced,frame style image=blueshade.png,
  opacityback=0.75,opacitybacktitle=0.25,
  colback=orange!5!white,colframe=orange!75!black,
  title=CodeFuse-CodeLLama-34B:]
\begin{lstlisting}[language=Go, breaklines=true]
    encrypted := make([]byte, len(s))
    for i, c := range s {
        if c >= 'a' && c <= 'z' {
            encrypted[i] = byte((c-'a'+2)%26 + 'a') // shift down two places
        } else if c >= 'A' && c <= 'Z' {
            encrypted[i] = byte((c-'A'+2)%26 + 'A')
        } else {
            encrypted[i] = byte(c) // non-alphabetic characters remain the same
        }
    }
    return string(encrypted)
}
\end{lstlisting}
\end{tcolorbox}

\subsection{Code Completion with Chinese prompts on CodeFuseEval-CN}

\begin{tcolorbox}
[colback=yellow!5!white,colframe=yellow!50!black,
  colbacktitle=yellow!75!black,title=Prompt:]
<|role\_start|>human<|role\_end|>\# language: Python \\
\includegraphics[width=\textwidth]{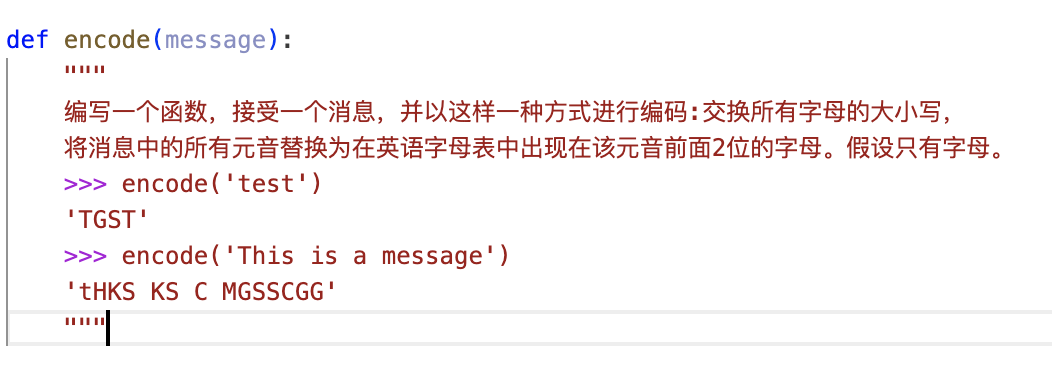}
<|role\_start|>bot<|role\_end|>
\end{tcolorbox}

\begin{tcolorbox}[enhanced,frame style image=blueshade.png,
  opacityback=0.75,opacitybacktitle=0.25,
  colback=orange!5!white,colframe=orange!75!black,
  title=CodeFuse-CodeLLama-34B:]
\begin{lstlisting}[language=Python, breaklines=true]
    encoded_message = ''
    vowels = {'a': 'y', 'A': 'Y', 'e': 'c', 'E': 'C', 'i': 'g', 'I': 'G', 'o': 'm', 'O': 'M', 'u': 's', 'U': 'S'}
    
    for letter in message:
        if letter.isalpha():
            if letter in vowels:
                encoded_message += vowels[letter]
            elif letter.islower():
                encoded_message += letter.upper()
            elif letter.isupper():
                encoded_message += letter.lower()
        else:
            encoded_message += letter

    return encoded_message
}
\end{lstlisting}
\end{tcolorbox}

\subsection{Text-to-Code on MBPP}

\begin{tcolorbox}
[colback=yellow!5!white,colframe=yellow!50!black,
  colbacktitle=yellow!75!black,title=Prompt:]
<|role\_start|>human<|role\_end|>
Write a function to find the number of subsequences having product smaller than k for the given non negative array. Your code should satisfy these tests:
\begin{lstlisting}
assert no_of_subsequences([1,2,3,4], 10) == 11 
assert no_of_subsequences([4,8,7,2], 50) == 9 
assert no_of_subsequences([5,6,7,8], 15) == 4 
\end{lstlisting}
<|role\_start|>bot<|role\_end|>
\end{tcolorbox}

\begin{tcolorbox}[enhanced,frame style image=blueshade.png,
  opacityback=0.75,opacitybacktitle=0.25,
  colback=orange!5!white,colframe=orange!75!black,
  title=CodeFuse-CodeLLama-34B:]
\begin{lstlisting}[language=Python, breaklines=true]
def no_of_subsequences(arr, k):
    n = len(arr)
    dp = [0] * n
    dp[0] = 1
    for i in range(1, n):
        dp[i] = dp[i-1]
        if arr[i] < k:
            dp[i] += 1
        for j in range(i):
            if arr[j] < k//arr[i]:
                dp[i] += dp[j]
    return dp[n-1]


import unittest

class TestNoOfSubsequences(unittest.TestCase):
    def test_no_of_subsequences(self):
        self.assertEqual(no_of_subsequences([1,2,3,4], 10), 111)
        self.assertEqual(no_of_subsequences([4,8,7,2], 50), 9)
        self.assertEqual(no_of_subsequences([5,6,7,8], 15), 4)


if __name__ == "__main__":
    unittest.main()
\end{lstlisting}
\end{tcolorbox}

\subsection{Data Science Analysis on DS-1000}

\begin{tcolorbox}
[colback=yellow!5!white,colframe=yellow!50!black,
  colbacktitle=yellow!75!black,title=Prompt:]
I'm using tensorflow 2.10.0.\\
I have a tensor of lengths in tensorflow, let's say it looks like this:
\begin{lstlisting}
[4, 3, 5, 2]
\end{lstlisting}

I wish to create a mask of 1s and 0s whose number of 1s correspond to the entries to this tensor, padded in front by 0s to a total length of 8. I.e. I want to create this tensor:
\begin{lstlisting}
[[0. 0. 0. 0. 1. 1. 1. 1.] 
[0. 0. 0. 0. 0. 1. 1. 1.] 
[0. 0. 0. 1. 1. 1. 1. 1.]
[0. 0. 0. 0. 0. 0. 1. 1.]]    
\end{lstlisting}

How might I do this? \\
A:\\
<code>\\
import tensorflow as tf\\\\\\
lengths = [4, 3, 5, 2]\\
</code>\\
result = ... \# put solution in this variable\\
BEGIN SOLUTION\\
<code>\\
\end{tcolorbox}

\begin{tcolorbox}[enhanced,frame style image=blueshade.png,
  opacityback=0.75,opacitybacktitle=0.25,
  colback=orange!5!white,colframe=orange!75!black,
  title=CodeFuse-CodeLLama-34B:]
\begin{lstlisting}[language=Python, breaklines=true]
max_length = tf.reduce_max(lengths)

# Create a tensor of 1s and 0s
mask = tf.sequence_mask(lengths, maxlen=max_length, dtype=tf.float32)

# Pad the mask with 0s to a total length of 8
padded_mask = tf.pad(mask, [[0, 0], [0, 8 - max_length]], constant_values=0)

result = padded_mask
\end{lstlisting}
\end{tcolorbox}

\subsection{Code Translation on CodeFuseEval-CodeTrans}

\begin{tcolorbox}
[colback=yellow!5!white,colframe=yellow!50!black,
  colbacktitle=yellow!75!black,title=Prompt:]
<|role\_start|>human<|role\_end|>
\begin{lstlisting}[language=Java, breaklines=true]
# Translate the following Java program to Python
# Java
import java.util.*;
import java.lang.*;

class Solution {
    public List<List<Integer>> getRow(List<List<Integer>> lst, int x) {
        List<List<Integer>> coords = new ArrayList<>();
        for (int i = 0; i < lst.size(); i++) {
            List<List<Integer>> row = new ArrayList<>();
            for (int j = lst.get(i).size() - 1; j >= 0; j--) {
                if (lst.get(i).get(j) == x) {
                    row.add(Arrays.asList(i, j));
                }
            }
            coords.addAll(row);
        }
        return coords;
    }
}
# Python
\end{lstlisting}
<|role\_start|>bot<|role\_end|>
\end{tcolorbox}

\begin{tcolorbox}[enhanced,frame style image=blueshade.png,
  opacityback=0.75,opacitybacktitle=0.25,
  colback=orange!5!white,colframe=orange!75!black,
  title=CodeFuse-CodeLLama-34B:]
\begin{lstlisting}[language=Python, breaklines=true]
def get_row(lst, x):
    coords = []
    for i in range(len(lst)):
        for j in range(len(lst[i])-1, -1, -1):
            if lst[i][j] == x:
                coords.append([i, j])
    return coords
\end{lstlisting}
\end{tcolorbox}

\subsection{Unit test case Generation on CodeFuseEval-UnitTest}

\begin{tcolorbox}
[colback=yellow!5!white,colframe=yellow!50!black,
  colbacktitle=yellow!75!black,title=Prompt:]
<|role\_start|>human<|role\_end|>
\begin{lstlisting}[language=Java, breaklines=true]
Write unit tests for the following code, without providing explanations: 
```
public class ExceptionUtils {
    private static final String[] CAUSE_METHOD_NAMES = {
        "getCause",
        "getNextException",
        "getTargetException",
        "getException",
        "getSourceException",
        "getRootCause",
        "getCausedByException",
        "getNested",
        "getLinkedException",
        "getNestedException",
        "getLinkedCause",
        "getThrowable",
    };

    public static Throwable getCause(final Throwable throwable, String[] methodNames) {
        if (throwable == null) {
            return null;
        }
        if (methodNames == null) {
            final Throwable cause = throwable.getCause();
            if (cause != null) {
                return cause;
            }
            methodNames = CAUSE_METHOD_NAMES;
        }
        return Stream.of(methodNames).map(m -> getCauseUsingMethodName(throwable, m)).filter(Objects::nonNull).findFirst().orElse(null);
    }
}
```
\end{lstlisting}
<|role\_start|>bot<|role\_end|> 
\\
\\
\\
\\
\\
\\
\\
\\
\\
\\
\\
\\
\\
\\
\\
\end{tcolorbox}

\begin{tcolorbox}[enhanced,frame style image=blueshade.png,
  opacityback=0.75,opacitybacktitle=0.25,
  colback=orange!5!white,colframe=orange!75!black,
  title=CodeFuse-CodeLLama-34B:]
\begin{lstlisting}[language=Java, breaklines=true]
```
import org.junit.Assert;
import org.junit.Test;

import java.util.Objects;
import java.util.stream.Stream;

public class ExceptionUtilsTest {
    @Test
    public void testGetCause_NullThrowable() {
        Throwable throwable = null;
        String[] methodNames = {"getCause"};
        Throwable output = ExceptionUtils.getCause(throwable, methodNames);
        Assert.assertEquals(null, output); 
    }
    @Test
    public void testGetCause_NullMethodNames() {
        Throwable throwable = new RuntimeException("test");
        String[] methodNames = null;
        Throwable output = ExceptionUtils.getCause(throwable, methodNames);
        Assert.assertEquals(null, output); 
    }
    @Test
    public void testGetCause_FoundCause() {
        Throwable throwable = new RuntimeException("test", new NullPointerException());
        String[] methodNames = {"getCause"};
        Throwable output = ExceptionUtils.getCause(throwable, methodNames);
        Assert.assertEquals(NullPointerException.class, output.getClass()); 
    }
    @Test
    public void testGetCause_NotFoundCause() {
        Throwable throwable = new RuntimeException("test");
        String[] methodNames = {"getCause"};
        Throwable output = ExceptionUtils.getCause(throwable, methodNames);
        Assert.assertEquals(null, output); 
    }
    @Test
    public void testGetCause_DefaultMethodNames_FoundCause() {
        Throwable throwable = new RuntimeException("test", new NullPointerException());
        Throwable output = ExceptionUtils.getCause(throwable);
        Assert.assertEquals(NullPointerException.class, output.getClass()); 
    }
    @Test
    public void testGetCause_DefaultMethodNames_NotFoundCause() {
        Throwable throwable = new RuntimeException("test");
        Throwable output = ExceptionUtils.getCause(throwable);
        Assert.assertEquals(null, output); 
    }
    private Throwable getCauseUsingMethodName(Throwable throwable, String methodName) {
        try {
            return (Throwable) throwable.getClass().getMethod(methodName).invoke(throwable);
        } catch (Exception e) {
            return null;
        }
    }
}
```
\end{lstlisting}
\end{tcolorbox}

\section{Evaluation results of CodeFuse-QWen-14B on NLP benchmark datasets}
\label{apx:codefuse_qwen_nlp_data}

\begin{table}[h]
  \caption{Comparisons of the performances of CodeFuse-QWen-14B, QWen-14B, and QWen-14b-chat on several NLP evaluation datasets. QWen-14B is a base model trained by Alibaba Cloud, QWen-14B-chat is a model fine-tuned by themselves on top of the QWen-14B base model, and CodeFuse-QWen-14B is a model fine-tuned by us using MFTCoder and code data.} 
  \centering
  \label{tab:app_qwen_nlp_comparation}
  \begin{tabular}{lc|ccc}
  \toprule
   & & \textbf{QWen-14B}  &  \textbf{QWen-14B-chat} & \textbf{CodeFuse-QWen-14B} \\
  \midrule
  \multirow{4}{*}{\textsc{Language}} & AFQMC & 69.00\% & 72.6\% & 71.99\% \\

  & CHID & 84.7\% & 72.3\% & 84.42\% \\
  & Wic & 50.9\% & 50.5\% & 55.02\% \\
  & WSC & 66.3\% & 66.3\% & 68.27\% \\
  & Average & 67.73\% &  65.43\% & \textbf{69.93\%} \\
  \midrule
  \multirow{6}{*}{\textsc{Reasoning}} & COPA & 93\% & 89\% & 95.00\% \\
  & CMNLI & 62.1\% & 60.3\% & 55.33\% \\
  & OCNLI & 58.2\% & 53.8\% & 49.22\% \\
  & AX-b & 49.5\% &58.5\% & 67.84\% \\
  & AX-g & 80.9\%  & 52.5\%  &  82.87\% \\
  & RTE & 71.5\%  &  51.6\% &  76.53\% \\
    & Average & 69.20\% & 60.95\% & \textbf{71.13\%} \\

  \midrule
  \multirow{3}{*}{\textsc{Understanding}} & CSL & 54.4\% & 55.6\% & 68.50\% \\
  & C3 & 90.8\% & 91.7\% & 91.01\% \\
  & EPRSTMT & 86.9\% & 91.2\% & 84.92\% \\
  & Average & 77.37\% & 79.50\% & \textbf{81.48\%} \\

   \midrule
  \multirow{3}{*}{\textsc{Examination}}
  & AX-b & 67.9\% & 66.4\% & 64.27\% \\
  & AX-g & 71.7\%  & 71.7\%  &  68.98\% \\
  & RTE & 84.4\%  & 80.3\% &  84.75\% \\
  & Average & \textbf{74.67\%} & 72.80\% & 72.67\% \\
  \bottomrule
  \end{tabular}
\end{table}

\end{document}